%% file: main.tex
\documentclass[a4paper,fleqn]{cas-sc}
\providecommand\SuspendTagging[1]{}
\providecommand\ResumeTagging[1]{}

\usepackage[authoryear]{natbib}
\usepackage{tabularx}
\usepackage{xcolor}
\usepackage{url}
\usepackage{algorithm}
\usepackage{algpseudocode}
\usepackage{tikz}
\usepackage{xltabular}
\usetikzlibrary{positioning,arrows.meta,shapes.geometric,backgrounds,fit,calc}

\newif\ifdraftmarks
\draftmarkstrue

\newcommand{\QS}{\mathit{QS}}
\newcommand{\Snet}{S_{\mathrm{net}}}
\newcommand{\Sbib}{S_{\mathrm{bib}}}
\newcommand{\Ssem}{S_{\mathrm{sem}}}
\newcommand{\Sproc}{S_{\mathrm{proc}}}
\newcommand{\file}[1]{\texttt{#1}}
\newcolumntype{Y}{>{\raggedright\arraybackslash}X}
\newcolumntype{P}[1]{>{\raggedright\arraybackslash}p{#1}}

\ExplSyntaxOn
\RenewDocumentCommand \printorcid { }
  {
    \seq_if_empty:NF \g_stm_orcid_seq
      {
        \group_begin:
          \tex_let:D \thefootnote \relax \footnotetext
            { \raggedright \textsc{orcid}(s):\c_space_token \seq_use:Nn \g_stm_orcid_seq { ;~ } }
        \group_end:
      }
  }
\ExplSyntaxOff

\ExplSyntaxOn
\cs_gset:Npn \__first_footerline:
  { \group_begin: \small \sffamily \__short_authors: \group_end: }
\ExplSyntaxOff

\begin{document}
\let\WriteBookmarks\relax
\def\floatpagepagefraction{1}
\def\textpagefraction{.001}

\shorttitle{PARIS: a multi-agent LLM framework for PRISMA-guided reviews}
\shortauthors{C. Cruz et al.}

\title [mode = title]{PARIS: A multi-agent LLM framework for end-to-end, PRISMA-guided systematic literature reviews with pre-synthesis corpus quality assessment}

\author[1]{Christophe Cruz}
\cormark[1]
\ead{christophe.cruz@u-bourgogne.fr}

\author[1,2]{Hussam Ghanem}

\author[1]{Samir Jabbar}

\author[1]{Fatiha Ait Kbir}

\author[1]{Hocine Cherifi}

\affiliation[1]{organization={Universit\'e Bourgogne Europe, CNRS, Laboratoire Interdisciplinaire Carnot de Bourgogne (ICB), UMR 6303},
            city={Dijon},
            postcode={21000},
            country={France}}

\affiliation[2]{organization={DAVI The Humanizers},
            city={Puteaux},
            country={France}}

\cortext[1]{Corresponding author}

\begin{abstract}
Systematic literature reviews (SRs) remain slow and labor-intensive, and most automation tools address single stages of the workflow. We present PARIS (PRISMA Agentic Research Intelligence System), an open-source framework in which a conductor agent running in Claude Code orchestrates 15 specialized large language model (LLM) agents, from scoping a research question to exporting a manuscript draft. Agents are specified declaratively, coordinate exclusively through persisted artifacts that the conductor validates before every transition, and delegate retrieval, bibliometric analysis and scoring to deterministic Python utilities. Three human checkpoints gate study inclusion, corpus adequacy and the synthesis outline. We also introduce QualityScore, a pre-synthesis index that aggregates normalized network, bibliometric, semantic and process sub-scores into a 0--100 grade. Two case studies illustrate both: AI chatbots and university students' learning (388 records, 100 included) and AI-assisted SR methods (131 records, 123 included, QualityScore 69.1 recomputed with a proxy network sub-score). They show end-to-end feasibility across domains and the diagnostic use of QualityScore: in the first run, corpus improvements chosen at the corpus-adequacy checkpoint raised the score from 74.6 to 81.9 and the number of pre-2023 studies from 1 to 19. Audits of both runs exposed failure modes: incomplete retrieval, counts that drifted between agents, unrecorded or delegated checkpoint decisions and a grade that depends on how citation data are handled. PARIS fully supports 8 of the 27 PRISMA 2020 items and partially supports 11. Output quality has so far been judged only by an internal reviewer agent; independent validation against human reviews is the next step.
\end{abstract}


\maketitle

\section{Introduction}\label{sec:intro}

\subsection{The systematic review bottleneck}\label{sec:intro:bottleneck}

Systematic literature reviews (SRs) are the reference method for synthesizing research evidence because they combine an explicit search, predefined eligibility criteria and a reproducible synthesis \citep{kitchenham2007guidelines,page2021prisma}. They are also expensive. Across 195 completed reviews registered in PROSPERO, the mean time from registration to publication was 67.3 weeks and the mean team comprised five authors \citep{borah2017analysis}; a formula-based estimate puts the personnel cost of a single review at about US\$141{,}000 \citep{michelson2019cost}. As publication volumes keep growing, the distance between knowledge production and knowledge synthesis widens unless parts of the review process are automated.

Automation of SRs has been studied for more than a decade \citep{tsafnat2014automation,omaraeves2015text,marshall2019toward,vandinter2021automation}. Most mature tools support one stage, typically title and abstract screening \citep{vandeschoot2021asreview,ouzzani2016rayyan}, and integrated toolchains operated by experienced teams have shown that a complete review can be produced in two weeks \citep{clark2020full}. Large language models (LLMs) have since been evaluated for screening \citep{khraisha2024llm,delgadochaves2025transforming}, data extraction \citep{gartlehner2024data} and risk-of-bias assessment \citep{lai2024assessing}. Two recent scoping reviews concluded that LLM support for reviews is ``on the rise, but not yet ready for use'' \citep{lieberum2025large} and that most work automates a single stage \citep{scherbakov2025emergence}.

\subsection{Problem statement}\label{sec:intro:problem}

We identify three problems that motivate this work.

\begin{enumerate}
\item \textbf{Stage fragmentation.} Errors propagate across stages (a record missed at retrieval cannot be recovered at synthesis), yet most systems are designed and evaluated stage by stage. LLM-agent pipelines that span the whole workflow have appeared only recently and mostly target clinical reviews with meta-analysis \citep{lin2026metapipe,huang2026lumen}; other agentic systems automate several stages under human oversight \citep{sami2024system,rouzrokh2025lattereview,wang2025trialmind,cao2025ottosr}.
\item \textbf{No corpus-level quality signal before synthesis.} Screening tools report progress or estimated recall, but no tool tells the reviewer whether the assembled corpus is structurally adequate (coherent, balanced, connected to the field's core literature) before the synthesis is written.
\item \textbf{Auditability of LLM pipelines.} Asked to reproduce the reference lists of 11 published SRs, GPT-3.5, GPT-4 and Bard hallucinated 39.6\%, 28.6\% and 91.4\% of the references they returned and recalled at most 13.7\% of the original references \citep{chelli2024hallucination}; LLM judges also favor their own outputs \citep{zheng2023judging,panickssery2024llm}. An automated review pipeline must therefore expose every intermediate decision for inspection and keep humans accountable for the consequential ones \citep{flemyng2025position,tourchimoghaddam2026arisma}.
\end{enumerate}

\subsection{Research questions and contributions}\label{sec:intro:contrib}

We address four research questions:
\textbf{RQ1} (architecture): how can an LLM multi-agent system execute the complete SR workflow while keeping every intermediate decision auditable and consequential decisions under human authority?
\textbf{RQ2} (corpus quality): can a composite index computed before synthesis give interpretable and actionable signals about corpus adequacy?
\textbf{RQ3} (feasibility): what do two end-to-end runs in different domains reveal about feasibility and failure modes?
\textbf{RQ4} (positioning): which PRISMA 2020 reporting items can such a system support, and how does it compare with existing tools and agentic systems?

The paper makes four contributions:
\begin{enumerate}
\item \textbf{The PARIS framework} (Section~\ref{sec:paris}): a conductor and 17 specialized agents (15 in the standard workflow, three of which run only when needed, an updater for living reviews and an optional retrofit utility), specified declaratively and coordinated through persisted, validated artifacts, with deterministic utilities for computation and three human checkpoints. The description is derived from the open-source implementation (release 0.6).
\item \textbf{QualityScore} (Section~\ref{sec:qs}): a formal definition of a four-dimension, pre-synthesis corpus quality index, its missing-data behavior, its diagnostic use and its limitations.
\item \textbf{Two case studies} (Sections~\ref{sec:setup}--\ref{sec:results}) on AI chatbots in higher education and on SR automation, including audits of both runs' artifacts and a structured analysis of the failure modes we observed.
\item \textbf{An item-level positioning} against the PRISMA 2020 checklist and a comparison with review platforms and agentic systems (Sections~\ref{sec:related:positioning} and \ref{sec:results:prisma}).
\end{enumerate}

We do not claim that PARIS outputs are equivalent to human-conducted SRs. The automated reviewer verdicts reported in Section~\ref{sec:results:review} are internal consistency checks, not evidence of manuscript quality.

The remainder of the paper is organized as follows. Section~\ref{sec:related} reviews related work. Section~\ref{sec:paris} describes the PARIS framework and Section~\ref{sec:qs} defines QualityScore. Section~\ref{sec:setup} presents the case-study setup and Section~\ref{sec:results} the results. Section~\ref{sec:discussion} discusses implications, responsible use, threats to validity and future work, and Section~\ref{sec:conclusion} concludes.

\section{Background and related work}\label{sec:related}

\subsection{Systematic review methodology and reporting}\label{sec:related:method}

PRISMA 2020 specifies 27 reporting items and a flow diagram for systematic reviews \citep{page2021prisma}, and its extension for scoping reviews adapts them to reviews that map a field rather than estimate effects \citep{tricco2018prismascr}. In computing, the guidelines of \citet{kitchenham2007guidelines} adapted SR methodology to software engineering. PRISMA is a \emph{reporting} guideline: compliance is a property of a report, not of a tool. We therefore describe PARIS as \emph{PRISMA-guided} and assess, item by item, which reporting elements its artifacts can support (Section~\ref{sec:results:prisma}). Guidance specific to AI in evidence synthesis is emerging. Cochrane, the Campbell Collaboration, JBI and the Collaboration for Environmental Evidence have issued a joint position statement \citep{flemyng2025position}, and the ARISMA guidelines treat AI as an ``inspected, benchmarked, logged, and reversible assistant rather than an autonomous reviewer'' \citep{tourchimoghaddam2026arisma}.

\subsection{Automation of individual review stages}\label{sec:related:stages}

\textit{Search and deduplication.} LLMs can draft Boolean queries for systematic searches \citep{wang2023chatgpt}, and cross-database deduplication has been automated by combining natural language processing with rules written by information specialists \citep{borissov2022deduklick}.

\textit{Screening.} Text mining for study identification was reviewed a decade ago \citep{omaraeves2015text}. Active-learning tools such as ASReview \citep{vandeschoot2021asreview} and review platforms such as Rayyan \citep{ouzzani2016rayyan} brought machine-learning prioritization to practitioners, and screening tools have been compared on their features and usability \citep{harrison2020software}. When to stop screening remains a central question; statistical stopping criteria trade additional effort against the risk of missed records \citep{callaghan2020statistical}. LLM-based screening has been evaluated with 18 LLMs across three reviews, which reduced one reviewer's workload by 33--93\% but depended strongly on how eligibility criteria were worded \citep{delgadochaves2025transforming}. A pre-registered evaluation of GPT-4 found human-level accuracy on some tasks that dropped once chance agreement and dataset imbalance were accounted for \citep{khraisha2024llm}.

\textit{Extraction and appraisal.} A proof-of-concept study compared Claude 2 with human extraction on 160 data elements from ten randomized trials \citep{gartlehner2024data}. RobotReviewer automated parts of risk-of-bias assessment for clinical trials \citep{marshall2016robotreviewer}, and LLMs reached 84.5--89.5\% correct domain-level judgments against expert assessments of 30 trials \citep{lai2024assessing}. Broader reviews compare the AI features of current review tools and the uses of LLMs across review stages \citep{bolanos2024artificial,lieberum2025large,scherbakov2025emergence}.

\subsection{LLM agents and multi-agent orchestration}\label{sec:related:agents}

LLM agents interleave reasoning with actions \citep{yao2023react}, and surveys map the design space of LLM-based autonomous agents \citep{wang2024survey}. Multi-agent frameworks such as MetaGPT encode standardized operating procedures as role specifications that exchange structured intermediate documents \citep{hong2024metagpt}. Practitioner guidance distinguishes \emph{workflows}, in which LLMs follow predefined paths, from \emph{agents}, which direct their own processes, and describes orchestrator--worker patterns in which a central model delegates subtasks \citep{anthropic2024building}. PARIS is built on Claude Code, in which subagents run in their own context window with a custom system prompt and restricted tool permissions \citep{anthropic2026subagents}. Agents that read retrieved documents are exposed to indirect prompt injection \citep{greshake2023indirect}.

\subsection{Agentic systems for literature review and synthesis}\label{sec:related:agentic}

Several systems generate literature syntheses or answer literature questions. AutoSurvey writes complete surveys through outline generation, parallel drafting and refinement \citep{wang2024autosurvey}, and PaperQA2 matched or exceeded subject-matter experts on literature search, summarization and contradiction detection tasks \citep{skarlinski2024paperqa2}.

Closer to PARIS are systems that automate SR stages. \citet{sami2024system} proposed a multi-agent system that generates search strings, filters titles, summarizes abstracts and analyzes the selected papers against research questions. LatteReview offers a provider-agnostic framework of modular agents for screening, relevance scoring and data extraction \citep{rouzrokh2025lattereview}. TrialMind supports search, screening and extraction for clinical evidence; on a benchmark of 100 published reviews (2,220 studies), its searches reached a recall of 0.711--0.834 against 0.138--0.232 for human-written baselines \citep{wang2025trialmind}. The otto-SR workflow automated screening, extraction and risk-of-bias assessment; it reported 96.7\% sensitivity and 97.9\% specificity in screening (human reviewers: 81.7\% and 98.1\%) and reproduced and updated an issue of 12 Cochrane reviews \citep{cao2025ottosr}. SLIDERS assembles evidence tables and reconciles conflicting excerpts with a code-writing agent, staying near 90\% accuracy on corpora of 6--11 million tokens \citep{joshi2026sliders}.

Two 2026 preprints are the closest to PARIS. meta-pipe is a ten-stage LLM-agent pipeline built on Claude that runs from literature search to statistical analysis, GRADE assessment and manuscript generation, with five mandatory human decision points; its authors present it as a system description without validation data \citep{lin2026metapipe}. LUMEN coordinates 11 LLM agents over six phases of a review with meta-analysis; it reports directional agreement with published meta-analyses on 13 outcomes and a median cost of US\$22.65 per review \citep{huang2026lumen}. Multi-stage benchmarks for evaluating such pipelines are also appearing \citep{zabaleta2026scilitbench}, and commercial platforms increasingly integrate AI assistance into screening and extraction \citep{covidence2026,rayyan2026,distillersr2026,eppireviewer2026,elicit2026}.

\subsection{Bibliometric foundations of corpus quality}\label{sec:related:biblio}

Bibliographic coupling links papers that share references \citep{kessler1963bibliographic}. Community structure in coupling networks is commonly measured by modularity \citep{newman2004finding} and detected with the Louvain algorithm \citep{blondel2008fast}, whose partitions can contain poorly connected communities, a problem addressed by the Leiden algorithm \citep{traag2019louvain}. The h-index summarizes the concentration of citation impact \citep{hirsch2005index}, and open scholarly infrastructures such as OpenAlex \citep{priem2022openalex} make these measures computable at no cost. They are routinely used to map research fields; to our knowledge, they have not been combined into a gate that decides whether a review corpus is ready for synthesis.

\subsection{Positioning}\label{sec:related:positioning}

Table~\ref{tab:systems} compares PARIS with review platforms and agentic systems along the dimensions that matter for this paper. Three observations frame our contribution. First, current platforms are broader than often assumed: according to their documentation, Covidence, Rayyan and DistillerSR support deduplication, screening, extraction, risk-of-bias assessment and PRISMA flow diagrams, and EPPI-Reviewer supports meta-analysis. Second, the strongest evidence for LLM automation comes from clinical systems evaluated against human reference standards \citep{wang2025trialmind,cao2025ottosr,huang2026lumen}. Third, end-to-end manuscript generation is no longer unique: meta-pipe shares PARIS's Claude-based agent architecture and mandatory human decision points \citep{lin2026metapipe}. PARIS differs in three respects. It targets domain-agnostic scoping and mapping reviews rather than clinical meta-analyses; it gates synthesis on a corpus-level quality index computed from bibliometric and semantic structure; and it coordinates agents exclusively through persisted artifacts that document every decision. Like meta-pipe, it has not yet been validated against human reviews, and unlike LUMEN it does not log costs.

\begin{table}[pos=tbp]
\caption{Comparison of PARIS with review platforms and agentic systems (feature claims for commercial platforms are taken from vendor documentation accessed on 16~October~2026).}\label{tab:systems}
\footnotesize
\begin{tabularx}{\textwidth}{@{}P{2.25cm}YYP{2.6cm}P{1.15cm}Y@{}}
\toprule
\textbf{System} & \textbf{Workflow scope} & \textbf{Automation approach} & \textbf{Human role} & \textbf{Corpus gate} & \textbf{Evidence reported}\\
\midrule
ASReview \citep{vandeschoot2021asreview} & Title/abstract screening & Active learning (prioritization) & Screens and decides when to stop & No & Simulations on labelled datasets\\
Rayyan \citep{ouzzani2016rayyan,rayyan2026} & Deduplication to PRISMA flow, incl.\ extraction and RoB & AI-assisted prioritization and extraction & Collaborative screening & No & Tool papers, user evaluations\\
Covidence \citep{covidence2026} & Deduplication, dual screening, full text, extraction, RoB, PRISMA flow & Review management & Dual independent reviewers & No & Tool evaluations \citep{harrison2020software}\\
DistillerSR \citep{distillersr2026} & Deduplication, screening, extraction, RoB, PRISMA reports & Generative-AI screening and extraction & Human-in-the-loop or automated & No & Vendor documentation\\
EPPI-Reviewer \citep{eppireviewer2026} & Screening, coding, meta-analysis, thematic synthesis & Text mining, LLM-based coding & Reviewer-driven & No & Vendor documentation\\
Elicit \citep{elicit2026} & Screening and extraction; partial search and reports & LLM & Reviewer verifies & No & Vendor documentation\\
Sami et al.\ \citep{sami2024system} & Search string to analysis by research question & Multi-agent LLM & Defines topic and questions & No & Small evaluation (software engineering)\\
LatteReview \citep{rouzrokh2025lattereview} & Screening, relevance scoring, extraction & Modular LLM agents, provider-agnostic & Configures review rounds & No & Framework evaluation (preprint)\\
TrialMind \citep{wang2025trialmind} & Search, screening, extraction (clinical) & LLM pipeline & Human--AI collaboration & No & 100-review benchmark; user study\\
otto-SR \citep{cao2025ottosr} & Screening, extraction, RoB, review updating & LLM workflow & Dual review of discrepancies & No & Benchmarks vs.\ human reviewers (preprint)\\
SLIDERS \citep{joshi2026sliders} & Evidence tables and reconciliation & LLM agents writing code & Follow-up questions & No & Three SR-style tasks (preprint)\\
meta-pipe \citep{lin2026metapipe} & Search to meta-analysis, GRADE and manuscript & Ten-stage Claude-based agent pipeline & Five mandatory decision points & No & System description, no validation (preprint)\\
LUMEN \citep{huang2026lumen} & Six phases of a review with meta-analysis & 11 LLM agents with model routing & Not reported & No & Agreement with published meta-analyses; cost logs (preprint)\\
AutoSurvey \citep{wang2024autosurvey} & Retrieval to written survey & LLM writing pipeline & Minimal & No & Text-quality evaluation\\
PaperQA2 \citep{skarlinski2024paperqa2} & Literature question answering with citations & Retrieval-augmented agent & Asks questions & No & Expert comparisons, benchmark\\
\textbf{PARIS} (this work) & Scoping, retrieval, deduplication, translation, screening, open-access full text, bibliometric, gap and TRL analysis, synthesis, internal review, export & Conductor, 15 LLM agents, deterministic utilities, artifact-mediated & Three checkpoints: inclusion, corpus quality, outline & Yes & Two case studies; internal reviewer only\\
\bottomrule
\end{tabularx}
\end{table}

\section{The PARIS framework}\label{sec:paris}

\subsection{Design principles}\label{sec:paris:principles}

Six principles, stated in the framework's conductor and agent specifications, shape PARIS.

\begin{enumerate}
\item \textbf{Workflow completeness before stage optimization.} PARIS covers the workflow from a one-sentence research goal to an exported manuscript, accepting that individual stages are less optimized than in specialized tools.
\item \textbf{Artifact-mediated coordination.} Every step persists its output as a named file in a run folder, and the conductor validates that the file exists and is well-formed before triggering the next step; it never proceeds ``on verbal confirmation alone''.
\item \textbf{Human authority at consequential decisions.} Three mandatory checkpoints stop the pipeline until the researcher decides: inclusion of borderline papers (A), adequacy of the corpus (B) and the synthesis outline (C).
\item \textbf{Quality before writing.} The corpus is measured (QualityScore, gaps, limitations) and presented to the researcher before any synthesis is written, with an explicit path to improve the corpus.
\item \textbf{Determinism where possible.} Retrieval, network analysis, scoring and plotting are implemented in Python utilities; LLMs are reserved for judgment and language tasks and are constrained by closed-world rules (citations only from the corpus, verbatim extraction, counts re-derived from decision logs).
\item \textbf{Publication formats as first-class outputs.} Word, LaTeX and slide exports are pipeline steps, not afterthoughts.
\end{enumerate}

\subsection{Architecture}\label{sec:paris:architecture}

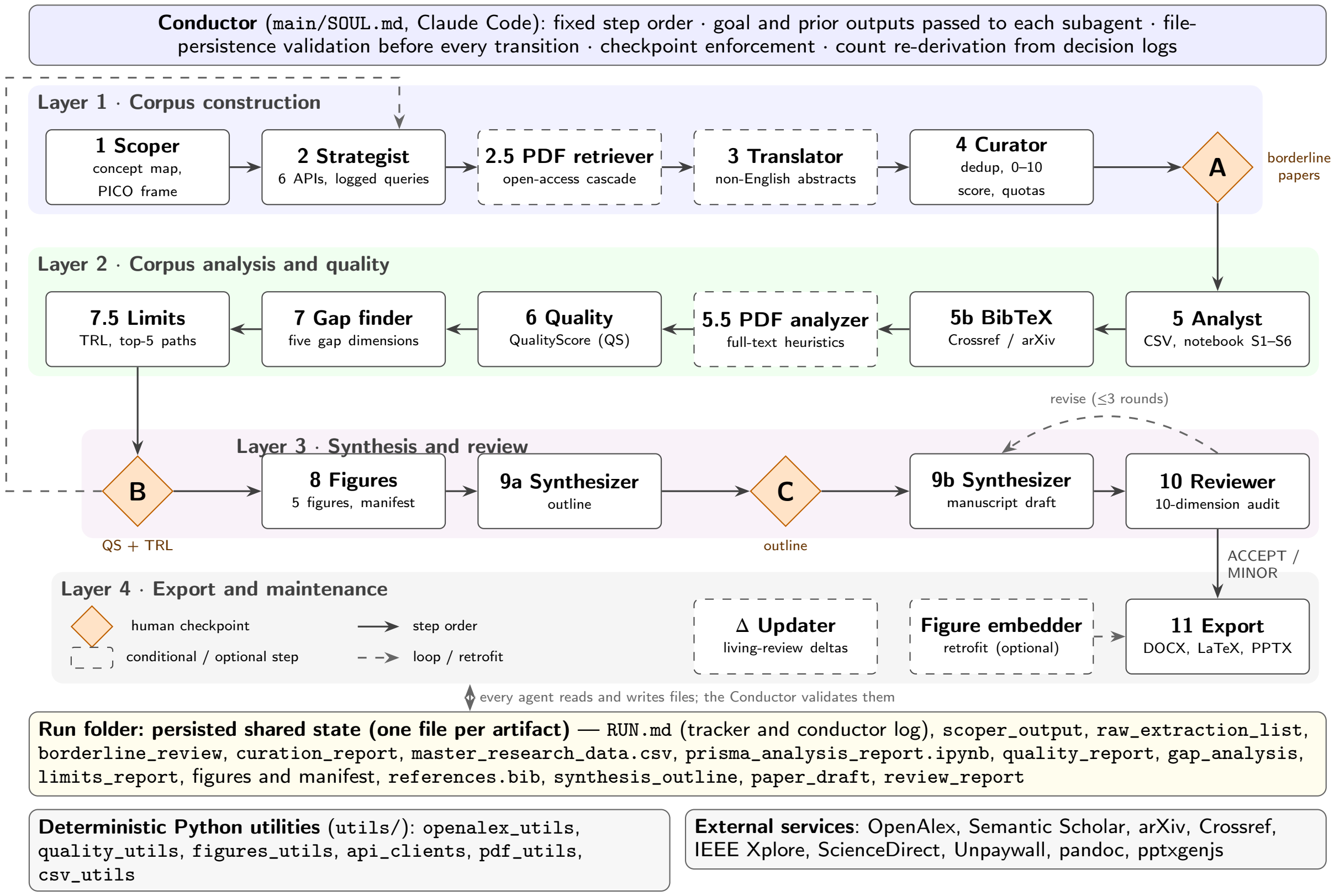
\begin{figure}[pos=tbp]
\centering
\resizebox{\linewidth}{!}{\input{figs/paris_architecture}}
\caption{Architecture of PARIS. The conductor forks one subagent per step and validates its persisted output before the next step. Diamonds are human checkpoints: A, inclusion of borderline papers; B, QualityScore and technology-readiness review, which may loop back to targeted delta queries; C, approval of the synthesis outline. Dashed boxes are conditional or optional steps. All agents communicate exclusively through files in the run folder; numerical work is delegated to deterministic Python utilities and external scholarly services.}\label{fig:architecture}
\end{figure}

Figure~\ref{fig:architecture} summarizes the architecture. In the taxonomy of \citet{anthropic2024building}, PARIS is a \emph{workflow} at the macro level, because the order of steps is fixed, whose steps are \emph{agents} at the micro level: each subagent decides how to accomplish its task, including writing and executing code. The conductor is specified in \file{main/SOUL.md}, whose first rule reads ``You orchestrate, you don't execute''. For each step it forks a subagent with the agent's specification, the primary research goal and the outputs of all previous steps; it then checks the expected output file against a persistence table, records the step in the run tracker (\file{RUN.md}) and either proceeds or holds a checkpoint. Agents are grouped in four layers: corpus construction, corpus analysis and quality, synthesis and review, and export and maintenance. The run folder is the only channel between agents, the audit trail of the review and the memory that lets a run resume across sessions.

\subsection{Declarative agent specifications}\label{sec:paris:specs}

Each agent is defined by a Markdown specification file (named \file{SOUL.md}) with a fixed structure: a \emph{mission}; the \emph{input} files to read; the \emph{output} files, given as templates with mandatory sections and tables; numbered \emph{behavior} rules; a one-line \emph{completion signal} that reports counts (for example, ``CURATOR CHECKPOINT --- borderline\_review.md written. N papers need human review. Waiting for Conductor decisions.''); and a short persona statement that sets the agent's stance (the scoper is ``a meticulous librarian''; the reviewer ``a demanding reviewer who has read all the supporting files''). Appendix~\ref{app:soul} reproduces an excerpt. Table~\ref{tab:agents} lists the agents. Eight Claude Code skills group the steps into working sessions that end at natural stopping points, typically a checkpoint: \file{/sota-new} (steps 1--4a), \file{/sota-curate} (4b--5.5), \file{/sota-quality} (6--8), \file{/sota-write} (9), \file{/sota-review} (10--11), \file{/sota-export} (11), \file{/sota-update} (delta runs) and the read-only audit \file{/sota-status}. This specification style resembles the standardized operating procedures of MetaGPT \citep{hong2024metagpt}, but PARIS agents exchange files rather than messages, so every intermediate result remains inspectable after the run.

\begin{table}[pos=tbp]
\caption{PARIS agents (release 0.6), grouped by layer, with their main artifacts. Steps 2.5, 3 and 5.5 are conditional; the figure embedder is an optional retrofit utility.}\label{tab:agents}
\footnotesize
\begin{tabularx}{\textwidth}{@{}P{0.6cm}P{2.7cm}YP{4.1cm}@{}}
\toprule
\textbf{Step} & \textbf{Agent} & \textbf{Function (from the agent specification)} & \textbf{Main artifact(s)}\\
\midrule
\multicolumn{4}{@{}l}{\textit{Layer 1 --- Corpus construction}}\\
1 & Scoper & 5--10 concepts; keyword map with synonyms and $\geq$1 exclusion term per concept; ambiguous terms flagged; scope boundaries; PICO frame & \file{scoper\_output.md}\\
2 & Strategist & API queries to six sources; key pre-flight check; rate limiting; pagination; verbatim query log; deduplication; contamination guard & \file{strategist\_output.md}, \file{raw\_extraction\_list.md}\\
2.5 & PDF retriever & Open-access full text through a five-source cascade & \file{pdfs/}, \file{pdf\_retrieval\_report.md}\\
3 & Translator & Language detection; abstract translation with confidence labels & \file{translation\_report.md}\\
4 & Curator & Deduplication; 0--10 relevance scores with rationale; cluster quotas; borderline protocol (checkpoint A) & \file{borderline\_review.md}, \file{curation\_report.md}\\
\midrule
\multicolumn{4}{@{}l}{\textit{Layer 2 --- Corpus analysis and quality}}\\
5 & Analyst & Master dataset; executed notebook (PRISMA flow, statistics, networks, keywords, trends, citation analysis) & \file{master\_research\_data.csv}, notebook\\
5b & BibTeX & Crossref BibTeX with arXiv fallback & \file{references.bib}\\
5.5 & PDF analyzer & Heuristic extraction of sections, limitations, future work, methods, open code and data & \file{pdf\_analysis\_report.md}\\
6 & Quality & QualityScore, bridge papers, recommendations & \file{quality\_report.md}\\
7 & Gap finder & Intersection, temporal, platform and methodological gaps; author-stated open questions & \file{gap\_analysis.md}\\
7.5 & Limits & Limitations per cluster; TRL matrix; reproducibility map; top-5 improvement paths (checkpoint B) & \file{limits\_report.md}\\
\midrule
\multicolumn{4}{@{}l}{\textit{Layer 3 --- Synthesis and review}}\\
8 & Figure & Five figures with captions and manifest & figures, \file{figures\_manifest.md}\\
9 & Synthesizer & Outline (checkpoint C), then full draft with closed-world citations & \file{synthesis\_outline.md}, \file{paper\_draft.md}\\
10 & Reviewer & Verification checklist; 10 scored dimensions; verdict; up to three revision rounds & \file{review\_report.md}\\
\midrule
\multicolumn{4}{@{}l}{\textit{Layer 4 --- Export and maintenance}}\\
11 & Export & DOCX (pandoc, citeproc), LaTeX (elsarticle), PPTX (13 slides) & \file{paper\_draft.docx}, \file{.tex}, \file{.pptx}\\
$\Delta$ & Updater & Delta retrieval with the original queries; append-only integration & \file{update\_log.md}, \file{delta\_report.md}\\
-- & Figure embedder & Retrofits drafts written before release 0.6 with embedded figures & regenerated exports\\
\bottomrule
\end{tabularx}
\end{table}

\subsection{Corpus construction}\label{sec:paris:construction}

\textit{Scoping.} The scoper decomposes the research goal into atomic concepts before expanding synonyms, requires at least one exclusion term per concept, marks terms whose meaning differs across subfields with a disambiguation rule, records in-scope, out-of-scope and edge cases, and states a PICO-style frame (population or domain, intervention or technique, comparison, outcome). It favors precision; recall is broadened later through Boolean disjunctions.

\textit{Retrieval.} The strategist converts the keyword map into queries for six sources in priority order: OpenAlex, Semantic Scholar, arXiv, Crossref, Elsevier ScienceDirect and IEEE Xplore. Before querying, it sends a one-result probe for each credentialed source and records authentication failures (HTTP 401/403) instead of skipping silently. It paginates fully, logs every query URL verbatim with the counts at each deduplication step, and applies the rate limits in Table~\ref{tab:apis}. On HTTP 429 it waits for the \texttt{Retry-After} period (60~s by default), retries once and otherwise skips and flags the source. A domain-contamination guard, added during development after polysemous terms had retrieved off-domain records, appends exclusion terms to queries that use polysemous concepts and scans the results with a regular expression before they are written.

\begin{table}[pos=tbp]
\caption{Source-specific rate limiting implemented by the strategist agent.}\label{tab:apis}
\footnotesize
\begin{tabularx}{\textwidth}{@{}P{3.2cm}P{4.4cm}Y@{}}
\toprule
\textbf{Source} & \textbf{Limit} & \textbf{Strategy}\\
\midrule
OpenAlex & Polite pool, 10 requests/s & Contact address (\texttt{mailto}) on every request; cursor pagination\\
Semantic Scholar & 1 request/s without key; higher quota with key & API key header when available\\
arXiv & 1 request every 3 s & 3-s pause between pages\\
Crossref & Polite pool, 50 requests/s & Contact address on every request\\
ScienceDirect, IEEE Xplore & Per-key quota & Honor HTTP 429 \texttt{Retry-After}; retry once, then skip and flag\\
\bottomrule
\end{tabularx}
\end{table}

\textit{Deduplication.} Records retrieved from several sources are merged with the key cascade of Algorithm~\ref{alg:dedup}, applied at retrieval and again by the curator before scoring. The arXiv identifier has priority because the same preprint can appear under different DOIs in arXiv and OpenAlex; a development run missed such a duplicate under a DOI-first rule. Duplicates are flagged, never deleted, and the record with the most complete metadata is kept.

\begin{algorithm}[t]
\caption{Three-key deduplication cascade (strategist and curator specifications).}\label{alg:dedup}
\footnotesize
\begin{algorithmic}[1]
\Require records $R$ from all sources
\Ensure retained records $U$, duplicate map $D$
\State $k_1(r)$: arXiv identifier from the DOI (\texttt{10.48550/arXiv.*}), URL or venue field
\State $k_2(r)$: DOI, lower-cased, without the \texttt{https://doi.org/} prefix
\State $k_3(r)$: title, lower-cased, punctuation removed, first 100 characters
\State $U \gets \emptyset$; $D \gets \emptyset$; index $I_j \gets \emptyset$ for $j=1,2,3$
\For{$r \in R$}
  \State $u \gets$ first indexed match $I_j[k_j(r)]$ for $j=1,2,3$ with $k_j(r)$ defined, else \textsc{none}
  \If{$u = \textsc{none}$}
    \State add $r$ to $U$ and to every $I_j$ for which $k_j(r)$ is defined
  \Else
    \State keep the more complete of $u$ and $r$ (abstract over none, DOI over none) in $U$ and the indexes
    \State $D[\text{discarded}] \gets \text{kept}$; flag the discarded record \texttt{[DUPE:source]}
  \EndIf
\EndFor
\State \Return $U$, $D$ \Comment{duplicates are reported, never deleted}
\end{algorithmic}
\end{algorithm}

\textit{Translation.} The translator detects non-English records from the title, abstract and venue for seven languages, tags them with ISO 639-1 codes and translates abstracts while preserving technical terms in the original language. Each translation carries a high, medium or low confidence label. No record is excluded because of its language.

\textit{Curation.} The curator scores every record from 0 (out of scope) to 10 (essential) with a one-sentence rationale and accepts records scoring at least 4 by default. Records scoring 3, or 4 when the abstract is ambiguous, form the borderline set, which is written to \file{borderline\_review.md} with the sentence that suggests relevance and a recommendation; the pipeline then stops at checkpoint~A until the researcher has decided on every record. To avoid a corpus dominated by one theme, any cluster holding more than 30\% of accepted papers is flagged, and accepted papers beyond the soft cap
\begin{equation}
N_{\max} = \left\lceil 1.5\, N_{\mathrm{acc}} / K \right\rceil
\label{eq:cap}
\end{equation}
($N_{\mathrm{acc}}$ accepted papers, $K$ concepts) are moved to a quota list that the conductor may override. The final counts are recomputed from the decision log rather than copied from the curator's summary line, a rule introduced after count errors were observed.

\subsection{Corpus analysis and quality}\label{sec:paris:analysis}

\textit{Analyst.} The analyst consolidates all screened records, with an inclusion flag, into \file{master\_research\_data.csv} and writes an executable notebook with six sections. These cover a PRISMA flow diagram whose counts are read from the curation report; descriptive statistics; a co-authorship network restricted to authors with at least two papers or degree three, partitioned by Louvain on its largest component with a fixed seed; keyword analysis (TF-IDF on up to 2,000 unigrams and bigrams and $k$-means clustering with $k=\min(6,K)$); temporal growth; and citation analysis from OpenAlex reference lists. The citation analysis builds a co-citation graph (edges with weight $\geq 2$) and a bibliographic-coupling graph (papers sharing $\geq 3$ references), ranks papers by PageRank, and identifies the five papers with the highest betweenness centrality in the coupling graph as \emph{bridge papers} that later agents must not omit. The notebook is executed in place, and the step cannot complete while any cell is unexecuted or has raised an error. When OpenAlex is unreachable, the citation section is skipped explicitly.

\textit{Bibliography and full text.} The BibTeX agent retrieves entries from the Crossref content-negotiation service, falls back to arXiv-style entries, and flags records with neither identifier. The PDF retriever tries, in order, the OpenAlex open-access URL, arXiv, the Elsevier full-text API (with institutional credentials), Unpaywall and Semantic Scholar. It pauses 0.5~s between requests, validates downloads by their \texttt{\%PDF} signature and never bypasses paywalls. The PDF analyzer (\file{pdf\_utils.py}, PyMuPDF) detects headings whose font size exceeds the body size by more than one point, or that are set in bold, and that match one of 19 section names. It extracts up to five limitation sentences from the limitations, discussion and conclusion sections, ranked by specificity (numbers and capitalized tokens), and up to three future-work sentences. It also detects framework and method names with regular expressions and sets open-code and open-data flags from repository and data-archive patterns. Extracted sentences must be verbatim.

\textit{Gaps and limitations.} The gap finder analyzes five dimensions. These are concept intersections (pairs addressed by 0--2 papers), temporal gaps (fewer than two papers per year in a cluster), platform or substrate gaps (fewer than three papers), methodological gaps (keyword-map methods with fewer than three papers) and open questions stated by the authors themselves. It must distinguish a genuine gap from records that were not retrieved and from emerging areas. The limits agent characterizes each cluster's shared assumptions, validation mode, scale and blind spots. It assigns each paper a technology readiness level (TRL, 1--9), adding one level each for a public benchmark, a named baseline and an independent reproduction. It also builds a reproducibility map and derives cross-cluster \emph{complementarity paths}, in which one cluster's strength addresses another's weakness, ranked by feasibility, impact and existing research mass into five improvement paths. When full text is available for fewer than 30\% of papers, it warns that TRL estimates rest on abstracts.

\textit{Checkpoint B.} After the limits agent, the conductor presents the QualityScore dashboard and the TRL summary together. If QualityScore is below 65, it recommends improving the corpus, and the researcher chooses to proceed or to improve. Improvement triggers targeted delta queries for the weak clusters, re-curation and re-analysis.

\subsection{Synthesis, review and export}\label{sec:paris:synthesis}

The figure agent produces five standard figures with colorblind-safe palettes at 300~dpi and a manifest of captions. The synthesizer first proposes an outline listing the clusters with their bridge papers, the section plan, figure placement and how each gap will be covered, and stops at checkpoint~C. After approval it writes the draft. Mandatory elements include highlights, keywords, an abstract of at most 250 words, a methodology section with actual counts and the QualityScore, one section per cluster that opens with its bridge papers and includes recent advances from the last 24 months, at least three cross-cutting themes, at least five future directions, a limitations section and the declarations required by Elsevier. Future directions are taken first from the limits agent's improvement paths and then from author-stated open questions, and are ``never invented''. A closed-world rule requires every cited paper to be present in the master dataset.

The reviewer agent reads the draft together with the curation report, the master dataset, the quality, gap and limits reports, the figure manifest and the bibliography. Before forming any judgment, it runs a verification checklist. The checklist confirms that every citation is present in the dataset, that PRISMA counts agree with the decision log, that the QualityScore is reported and bridge papers are cited, that future directions trace to the gap and limits reports, that figure counts match the manifest, and that the journal elements are present. It then scores ten dimensions from 1 to 5, weighted as critical, high, medium or low, and issues a verdict (accept, minor revision, major revision or reject). Hallucinated citations and count inconsistencies are always critical. Major issues trigger up to three revision rounds, keeping each previous draft. On acceptance or minor revision, the export agent produces a Word file with pandoc and citeproc, a LaTeX source in the \texttt{elsarticle} class and a slide deck. It validates the deck and searches the outputs for leftover placeholders. The updater refreshes a completed run by re-executing the original queries for the period since the extraction date, deduplicating by DOI against the existing dataset, scoring with the same criteria and appending, never overwriting, the accepted papers.

\subsection{Integrity safeguards and implementation}\label{sec:paris:implementation}

Several safeguards recur across the specifications: persistence gates before every transition; counts re-derived from decision logs; closed-world citations checked against the dataset and the bibliography; verbatim extraction; flagging instead of deleting duplicates and quota decisions; append-only updates with versioned drafts; explicit reporting of API failures; and the three human checkpoints. Table~\ref{tab:executors} makes explicit which parts of the workflow are executed by LLM agents, which by deterministic code and how each is verified. This separation matters for reproducibility: the deterministic parts can be re-run exactly, whereas LLM decisions can only be audited through the artifacts they leave.

\begin{table}[pos=tbp]
\caption{Executors, determinism and verification of the main functions in PARIS release 0.6. ``LLM'' denotes a Claude Code subagent following its specification; such agents may write and execute code at run time.}\label{tab:executors}
\footnotesize
\begin{tabularx}{\textwidth}{@{}P{3.6cm}P{3.7cm}P{2.3cm}Y@{}}
\toprule
\textbf{Function} & \textbf{Executor} & \textbf{Determinism} & \textbf{Verification}\\
\midrule
Scoping and query formulation & LLM (scoper, strategist) & Non-deterministic & Queries logged verbatim with counts\\
Retrieval and pagination & LLM issuing API calls & Depends on API state & Key pre-flight, failure flags, per-source counts\\
Deduplication (Algorithm~\ref{alg:dedup}) & LLM following the specification & Non-deterministic & Duplicate counts and flags kept\\
Screening, clustering, quotas & LLM (curator) & Non-deterministic & Decision log with rationale; checkpoint A\\
Open-access full-text retrieval & Python (\file{api\_clients.py}) & Deterministic\textsuperscript{a} & Signature check; retrieval report\\
Full-text extraction & Python (\file{pdf\_utils.py}) & Deterministic & Verbatim-sentence rule\\
Corpus analytics & LLM-written notebook; \file{openalex\_utils.py} & Mostly deterministic & Executed notebook; error checks\\
QualityScore & Python (\file{quality\_utils.py}) & Deterministic except Louvain\textsuperscript{b} & Component values printed; checkpoint B\\
Gap and limits analysis & LLM (gap finder, limits) & Non-deterministic & ``Every claim cites a paper count''\\
Figures & Python (\file{figures\_utils.py}) or LLM-written code & Deterministic & Manifest; reviewer dimension 10\\
Synthesis & LLM (synthesizer) & Non-deterministic & Closed-world citations; checkpoint C\\
Review & LLM (reviewer) & Non-deterministic & Checklist against dataset and decision log\\
Bibliography and export & Crossref API; pandoc; pptxgenjs & Deterministic & Deck validation; placeholder search\\
\bottomrule
\end{tabularx}
\par\vspace{2pt}{\footnotesize \textsuperscript{a}Given the state of the external services. \textsuperscript{b}The Louvain partition in \file{quality\_utils.py} is not seeded (Section~\ref{sec:qs:limits}).}
\end{table}

PARIS release 0.6 consists of the conductor and 17 agent specifications, eight Claude Code skills and six Python modules (\file{openalex\_utils}, \file{quality\_utils}, \file{figures\_utils}, \file{api\_clients}, \file{pdf\_utils} and \file{csv\_utils}). It relies on pandas, NetworkX, python-louvain, scikit-learn, Matplotlib and PyMuPDF, on Jupyter for notebook execution, and on pandoc and pptxgenjs for export; it is distributed under the MIT licence \citep{cruz2026paris}. The framework evolved through six releases. Release 0.2 introduced API retrieval, translation, gap analysis, figures, updates, cluster balancing and the first checkpoints; 0.3 added open-access full text and the limits agent; 0.4 added export; 0.5 added the arXiv-first deduplication key, rate limiting, API pre-flight checks, the normalized QualityScore with grades, the BibTeX agent, the shared utilities and the contamination guard; 0.6 added embedded figures.

\section{QualityScore: a pre-synthesis corpus quality index}\label{sec:qs}

\subsection{Requirements}\label{sec:qs:req}

QualityScore is designed to be computable before synthesis from artifacts that the pipeline already produces; to be decomposable, so that a low value points to a corrective action; to combine sub-scores normalized to $[0,1]$; to apply in any domain; and to be transparent, with the implementation printing every component. It measures the \emph{structural adequacy} of a corpus, not the recall of the search: a corpus can score well while missing relevant studies.

\subsection{Definition}\label{sec:qs:def}

Let $A$ be the set of $N$ accepted papers and $n_{\mathrm{raw}}$ the number of screened records. The composite score is
\begin{equation}
\QS = 100\left(0.30\,\Snet + 0.25\,\Sbib + 0.20\,\Ssem + 0.25\,\Sproc\right),
\label{eq:qs}
\end{equation}
with grades A ($\QS \geq 80$), B ($65 \leq \QS < 80$), C ($50 \leq \QS < 65$) and D ($\QS < 50$). The four sub-scores, implemented in \file{utils/quality\_utils.py}, are defined as follows; Table~\ref{tab:qscomponents} summarizes them with their typical corrective actions.

\textit{Network.} Let $G=(V,E)$ be the bibliographic-coupling graph of the accepted papers resolved in OpenAlex, with an edge between papers $i$ and $j$ when their reference lists $R_i$ and $R_j$ share at least $\theta$ works, weighted by $|R_i \cap R_j|$ ($\theta=2$ in the utility default; $\theta=3$ in the analyst notebook). With $Q$ the modularity of the Louvain partition of $G$ \citep{blondel2008fast}, or of a greedy modularity partition when the Louvain package is unavailable,
\begin{equation}
\Snet = \min\!\left(Q/0.7,\; 1\right),
\label{eq:snet}
\end{equation}
so that modularity values of 0.7 and above saturate. Modularity above about 0.3 is commonly read as significant community structure \citep{newman2004finding}.

\textit{Bibliometric.} With citation counts $c_1 \geq c_2 \geq \dots \geq c_N$ (enriched from OpenAlex), let $h = \max\{k : c_k \geq k\}$ be the corpus h-index \citep{hirsch2005index}. Let $H_t = -\sum_y p_y \ln p_y / \ln|Y|$ be the Shannon entropy of the publication-year distribution normalized by the number $|Y|$ of distinct years. Let $G_v = \sum_{i<j}|x_i-x_j| / (m^2 \bar{x})$ be the Gini coefficient of the counts $x_1,\dots,x_m$ of papers per venue, and $f$ the fraction of papers with more than 50 citations. Then
\begin{equation}
\Sbib = \tfrac{1}{4}\left[\min\!\left(\tfrac{h}{30},1\right) + H_t + (1 - G_v) + \min\!\left(\tfrac{f}{0.15},1\right)\right].
\label{eq:sbib}
\end{equation}

\textit{Semantic.} Abstracts are represented by TF-IDF vectors (at most 1,000 features, minimum document frequency 2, English stop words removed). Let $\bar{s}$ be the mean cosine similarity between cluster centroids and $\rho$ the mean pairwise cosine similarity within a random sample of $\min(200,N)$ abstracts (fixed seed). Let $\kappa$ be the fraction of keyword-map terms longer than three characters that occur at least twice in the concatenated accepted abstracts. Then
\begin{equation}
\Ssem = \tfrac{1}{3}\left[(1-\bar{s}) + \kappa + (1-\rho)\right].
\label{eq:ssem}
\end{equation}

\textit{Process.} With the inclusion rate $r = N/n_{\mathrm{raw}}$ and
\begin{equation}
\varphi(r) =
\begin{cases}
r/0.30 & r < 0.30,\\
1 & 0.30 \leq r \leq 0.60,\\
\max\!\left(0,\, 1-(r-0.60)/0.40\right) & r > 0.60,
\end{cases}
\label{eq:phi}
\end{equation}
the Gini coefficient $G_s$ of papers per source database (computed on records retrieved from a single source) and the mean curator relevance score $\bar{m}$ on the 0--10 scale,
\begin{equation}
\Sproc = \tfrac{1}{3}\left[\varphi(r) + (1-G_s) + \min\!\left(\bar{m}/8,1\right)\right].
\label{eq:sproc}
\end{equation}

\begin{table}[pos=tbp]
\caption{QualityScore sub-scores, their components and typical corrective actions.}\label{tab:qscomponents}
\footnotesize
\begin{tabularx}{\textwidth}{@{}P{1.9cm}P{0.9cm}P{5.4cm}Y@{}}
\toprule
\textbf{Sub-score} & \textbf{Weight} & \textbf{Components (saturation)} & \textbf{A low value suggests \dots}\\
\midrule
$\Snet$ & 0.30 & Louvain modularity of the bibliographic-coupling graph (0.7); coverage penalty below 40\% OpenAlex coverage & fragmented or poorly resolved reference structure: add identifiers, query under-represented communities, check cluster definitions\\
$\Sbib$ & 0.25 & Corpus h-index (30); normalized year entropy; $1-$venue Gini; share of papers with $>$50 citations (15\%) & missing seminal work or skewed coverage: backward and forward snowballing, wider date range, more sources\\
$\Ssem$ & 0.20 & $1-$inter-cluster similarity; keyword coverage; $1-$redundancy & vocabulary drift, near-duplicate content or overlapping clusters: revise the keyword map, merge or split clusters\\
$\Sproc$ & 0.25 & Inclusion-rate band (30--60\%); $1-$source Gini; mean relevance (8/10) & imprecise or overly narrow search, single-source dominance: rebalance queries and sources\\
\bottomrule
\end{tabularx}
\end{table}

\subsection{Missing citation data}\label{sec:qs:missing}

$\Snet$ requires OpenAlex reference lists. The quality agent's specification applies a coverage penalty when fewer than 40\% of accepted papers are resolved,
\begin{equation}
\Snet \leftarrow \Snet \cdot c/0.40 \qquad (c < 0.40),
\label{eq:penalty}
\end{equation}
where $c$ is the resolved fraction, and it flags the network sub-score as a lower bound. When OpenAlex is unavailable for more than 60\% of the corpus, the specification asks for $\Sbib$, $\Ssem$ and $\Sproc$ only, and the corresponding skill asks the conductor to report how weights were redistributed. The specification does not fix a single rule for forming the composite in this case. Section~\ref{sec:results:qs} shows that the choice changes the grade. We therefore recommend reporting $\Snet$ as missing and giving the interval $[\QS(\Snet{=}0),\,\QS(\Snet{=}1)]$ alongside any point estimate.

\subsection{Properties and limitations}\label{sec:qs:limits}

The weights and saturation points (0.7, 30, 0.15, the 30--60\% band, 8/10) were set by the authors and have not been calibrated against expert judgments. Several properties of the implementation also matter when interpreting values.
\begin{itemize}
\item \textit{Network.} The Louvain call in \file{quality\_utils.py} is not seeded, so $\Snet$ can vary between runs on the same graph (it was seeded by a local change during Run~1; Section~\ref{sec:results:qs}). Modularity depends on graph density, and Louvain can return poorly connected communities, a weakness the Leiden algorithm corrects \citep{traag2019louvain}. The coupling threshold differs between the notebook ($\theta=3$) and the utility default ($\theta=2$) and must be reported. Papers without coupling edges do not enter $G$. The corpus-level utility fetches references by DOI only (up to 400 papers), so arXiv-only papers contribute no references.
\item \textit{Bibliometric.} Raw citation counts are not normalized by field or age, which penalizes young fields, and the Gini coefficient is unstable when there are few venues. Venue names are truncated to 50 characters. Normalizing the year entropy by the number of distinct years rewards even spread over a few years as much as over many.
\item \textit{Semantic.} Keyword coverage depends on the format of the scoper's keyword table, which is parsed with a regular expression, and on abstracts only. When the table does not match the expression, the utility silently sets coverage to 0.5; in Run~1 the quality agent reformatted its input to avoid this.
\item \textit{Process.} The inclusion-rate band penalizes precise searches and rewards broad ones, and $\bar{m}$ depends on how the LLM curator calibrates its scores.
\end{itemize}

\section{Case studies: setup}\label{sec:setup}

\subsection{Protocol}\label{sec:setup:protocol}

Each run was initialized from a run template (\file{RUN.md}) stating the research goal, domain, date range, target databases and available API credentials. The two runs differ in their execution environment. Run~2 used Claude Sonnet 4.6 for the conductor and all subagents, with the standard start prompt (``Act as the PRISMA Conductor \dots{} start a new run from \file{RUN.md}''), and PARIS releases 0.4--0.5 (13--14~October~2026). Run~1 used Claude Opus 5.5 for the conductor and all subagents, in Claude Code (Visual Studio Code extension, Windows), with the initial public release 0.5 of PARIS cloned on 14~October, which lacks the figure embedder of release 0.6. It was started with the \file{/sota-new} skill and continued with \file{/sota-curate}, \file{/sota-quality}, \file{/sota-write} and \file{/sota-review} between 14 and 26~October~2026. In Run~1 the conductor forked each agent as a subagent with a prompt that it wrote from the agent's specification, the run folder and the research goal; these prompts added run-specific constraints that the specifications do not contain, such as no contact address in API requests, citation keys equal to paper identifiers, image syntax for figures and scripted self-checks. Human input was limited to the checkpoint decisions, which members of the author team made or, in Run~1, delegated (Section~\ref{sec:setup:run1}). We report process counts (records, borderline decisions, inclusions), coverage (OpenAlex resolution, open-access full text), QualityScore and its components, the outputs of the gap and limits agents, manuscript characteristics, internal reviewer verdicts and observed failures. For both runs we audited the complete run folder: query logs, decision logs, dataset, notebook, quality, gap and limits reports, figure manifest, review report and the conductor's log in \file{RUN.md}.

\subsection{Run 1: AI chatbots and university students' learning}\label{sec:setup:run1}

The goal was to review empirical evidence on how AI chatbots (rule-based, retrieval-based and LLM-based conversational agents such as ChatGPT) affect the learning outcomes, engagement, self-regulation and academic integrity of university students (2018--2026; educational technology, cs.CY, cs.HC, cs.CL). No API key was available and the researcher declined to supply a contact address, so OpenAlex and Crossref were queried without the \texttt{mailto} parameter of Table~\ref{tab:apis}, Elsevier and IEEE Xplore were not queried, and Semantic Scholar returned HTTP~429 on every query. The scoper defined eight concepts and four outcome sub-queries (learning outcomes, engagement and motivation, self-regulation and metacognition, academic integrity). In a first round, the strategist captured only the first relevance-ranked page of each query, to respect the run template's ``about 200 papers'' parameter, while logging the reported totals (18,336--94,521 OpenAlex hits per query, inflated by full-text matching). Its 240 records (OpenAlex 140, arXiv 60, Crossref 40) yielded 192 unique records. The PDF retriever ran at its specified position, before curation, on all 192 records, although its specification asks it to restrict downloads to accepted papers, which do not exist at that point; it obtained 121 open-access PDFs (63.0\%). The translator handled six non-English records; its first execution, interrupted by the user, had already rewritten the extraction list without writing its report. The curator accepted 58 records, rejected 86 and flagged 48 as borderline, and the conductor's recount of the decision log matched.

At checkpoint~A, the researcher delegated the decision to the conductor (``include the papers that can be included and exclude the others'', translated from French). The conductor retrieved the full abstracts of 41 of the 48 records from OpenAlex, read two PDFs and wrote an inclusion rule: empirical data from higher-education students using a chatbot, with a learning, engagement, self-regulation or integrity outcome; usage-, attitude- or acceptance-only, non-higher-education, conceptual and unverifiable records excluded. It recorded a decision and a reason for every record in \file{RUN.md}: 21 included and 27 excluded, departing from the curator's recommendation in 12 cases. The first-pass corpus held 79 papers. Its learning-outcome cluster (32 papers) exceeded the soft cap of Eq.~\eqref{eq:cap} (30), and the conductor overrode the quota so that no eligible study was excluded for balance.

The quality agent graded this corpus 74.6 (B) and reported that only one accepted paper predated 2023, that no rule-based or retrieval-based chatbot was covered and that nine works cited by three or four corpus papers were missing. At checkpoint~B the researcher chose to improve the corpus. Because the updater only queries publications newer than the extraction date, the conductor implemented the improvement through subagents that followed the strategist's and updater's specifications: backward snowballing of ten OpenAlex works, and a paginated title-and-abstract query for 2018--2022 and non-LLM chatbots on OpenAlex, Crossref and arXiv. This second round identified 248 records, 196 of them new after deduplication against the first round. Full-text retrieval (60 PDFs, 30.6\%; most missing texts were paywalled publisher articles without an open-access copy), translation and curation (20 accepted, 152 rejected, 24 borderline) followed. The researcher again delegated checkpoint~A (``decide yourself'', translated from French). The conductor included four and excluded 20 borderline records and, after rechecking six first-round papers flagged during full-text analysis, excluded three of them post hoc (survey data described by their authors as illustrative, three participants, chatbot use inferred rather than observed). The combined PRISMA 2020 flow has two identification arms: 488 records identified (240 and 238 from databases, 10 from citation searching), 100 duplicates removed, 388 screened, 238 excluded on title and abstract, 150 assessed for eligibility, 50 excluded and 100 studies included (76 from the first database round, 17 from the second and 7 from citation searching). The quality agent then graded the corpus 81.9 (A), and the researcher chose to proceed.

The four outcome clusters hold 46, 13, 23 and 18 papers. Nineteen studies were published before 2023, and 11 explicitly test rule-based (9) or retrieval-based (2) chatbots. Open-access full text was available for 56 of the 100 included studies. At checkpoint~C the researcher approved the outline as proposed. The run's output is summarized in Section~\ref{sec:results:chatbots}.

\subsection{Run 2: AI-assisted systematic review methods}\label{sec:setup:run2}

The run template stated the goal as positioning PARIS among AI and NLP methods for automating systematic reviews (tools, screening and synthesis methods, knowledge graphs) and identifying the gaps it fills (2018--2026; cs.IR, cs.CL, cs.AI, cs.DL). The strategist ran 25 OpenAlex, six arXiv and one Crossref query. Semantic Scholar returned HTTP 429 without an API key and was skipped, and IEEE Xplore was skipped because its key was not in the environment. Only the first five or six records of most OpenAlex queries were captured, because the agent fetched result pages with a web-fetch tool whose responses were truncated instead of paginating the API, and off-topic records were discarded during retrieval without being logged. For the same reason, the extraction list stores, in place of abstracts, one-sentence snippets of 43--109 characters (median 68) that paraphrase the fetched pages; the dataset's abstract field holds these snippets, so curation and every later abstract-based analysis operated on them. About 285 captured records yielded about 137 unique records, among which the curator confirmed 15 duplicates. Its decision log lists 133 scored records: 125 accepted, seven borderline records all excluded at checkpoint~A, and one rejected as off-domain. The researcher overrode the curator's recommendation to include three of the seven borderline records. The dataset built by the analyst, on which all later steps operate, contains 131 records and 123 included papers (76 retrieved through OpenAlex, 46 through arXiv, one through Crossref): two accepted records and one excluded borderline record were lost, and one duplicate was added, during consolidation. The analyst did not write the executed notebook that its specification requires; the run tracker marks it as pending, and later agents computed their analyses in ad hoc code. Full text was retrieved before curation, from arXiv only and for 30 of the 60 arXiv records, selected by citation count; 29 of these papers were included (24\%). No delta cycle was run. OpenAlex was unreachable during the quality step, so $\Snet$ could not be computed; the quality agent substituted a proxy value of 0.55 based on TF-IDF cluster separation, which the specification does not define, and selected bridge papers by text similarity instead of coupling. The run tracker records the outline as approved by the user at checkpoint~C, but the outline file still carries the status ``awaiting conductor approval'' and the run log records no decision. The resulting review of AI-assisted SR methods is reported in Section~\ref{sec:results:landscape} as an output of the case study; it is not the basis of this paper's related work.

\section{Results}\label{sec:results}

\subsection{Corpus construction}\label{sec:results:corpus}

Table~\ref{tab:corpus} summarizes both runs. The inclusion rates differ sharply (25.8\% vs.\ 93.9\%). Run~2's queries targeted a narrow topic, and its curator scored included papers 6.98 on average on the 0--10 scale, with two thirds at 7 or above. Its inclusion rate, however, overstates the precision of the search, because off-topic records were discarded during retrieval without being counted. Run~1's inclusion rate fell from 41.1\% in the first round to 25.8\% after the second, whose Crossref queries could not express Boolean conjunctions and returned many records without any chatbot term; its included papers scored 6.28 on average. In both runs the borderline sets were small compared with the number of screened records (72 of 388 and 7 of 131). This is the intended benefit of LLM curation, but it also means that one LLM screener made all non-borderline decisions, unlike the dual independent screening recommended for SRs. In Run~1, the borderline decisions themselves were delegated to the conductor model.

\begin{table}[pos=tbp]
\caption{Corpus characteristics of the two case studies, verified against their run folders. Superscripts refer to the notes.}\label{tab:corpus}
\footnotesize
\begin{tabularx}{\textwidth}{@{}P{5.0cm}YY@{}}
\toprule
\textbf{Characteristic} & \textbf{Run 1: AI chatbots in higher education} & \textbf{Run 2: AI-assisted SR methods}\\
\midrule
PARIS release & 0.5 (initial release) & 0.4--0.5\\
Sources returning records & 3 of 4 attempted (OpenAlex, arXiv, Crossref) & 3 of 5 attempted (OpenAlex, arXiv, Crossref)\\
Unique records screened & 388 (192 + 196)\textsuperscript{b} & 131\textsuperscript{a}\\
Borderline decisions at checkpoint A (included/excluded) & 72 (25/47), delegated to the conductor; 6 rechecks (3 excluded) & 7 (0/7)\\
Included after first pass & 79 & 123\textsuperscript{a}\\
Delta cycle & +17 targeted (2018--2022, non-LLM); +7 snowballing; $-3$ post hoc & none\\
Final corpus & 100 & 123\\
Inclusion rate & 25.8\% & 93.9\%\\
Thematic clusters & 4 & 10\\
Open-access full text & 56/100 (56\%); 181 of 388 screened records & 29/123 (24\%), arXiv only, retrieved before curation\\
OpenAlex coverage at quality step & 100/100 resolved; 58/100 with reference lists & 0 (service unreachable)\\
Corpus h-index & 35 (23 before the delta cycle) & 29\textsuperscript{c}\\
Papers with $>$50 citations & 29/100 (29.0\%) & 20/123 (16.3\%)\textsuperscript{c}\\
Manuscript produced & $\approx$9,500 words, 100 cited works, 6 figures & $\approx$8,650 words, 28 cited works, 5 figures\\
Internal reviewer verdict & minor revision (0 critical, 14 minor issues) & accept with minor revisions\\
\bottomrule
\end{tabularx}
\par\vspace{2pt}{\footnotesize \textsuperscript{a}Dataset used by all later steps; the curation decision log lists 133 scored records and 125 accepted (Section~\ref{sec:setup:run2}). \textsuperscript{b}Two retrieval rounds; four second-round records already present in the first round are counted as duplicates. \textsuperscript{c}Citation counts recorded at retrieval; OpenAlex enrichment was unavailable.}
\end{table}

\subsection{QualityScore}\label{sec:results:qs}

Table~\ref{tab:qsresults} reports the sub-scores and their weighted contributions. Applying Eq.~\eqref{eq:qs} to the reported sub-scores reproduces all composite scores. For both runs, the run folders also allow the sub-scores to be checked against Eqs.~\eqref{eq:snet}--\eqref{eq:sproc}; Run~1's were computed with \file{quality\_utils.py} in the executed notebook.

\begin{table}[pos=tbp]
\caption{QualityScore results. Contributions are $100\,w_i S_i$. For Run~1, the upper block gives the final corpus (100 papers), after the corpus-improvement cycle triggered at checkpoint~B; the middle blocks give the corpus before that cycle (79 papers) and the sensitivity of the final score to the network computation. Run~1's final contributions are rounded; computed from unrounded sub-scores, they sum to 81.94. For Run~2, $\Snet$ is an ad hoc proxy; ``recomputed'' applies Eq.~\eqref{eq:phi} to the inclusion-rate term, which the run had computed differently. The lower block shows the Run~2 composite under the missing-data rules of Section~\ref{sec:qs:missing}.}\label{tab:qsresults}
\footnotesize
\begin{tabularx}{\textwidth}{@{}P{4.0cm}P{0.9cm}YYY@{}}
\toprule
\textbf{Sub-score} & \textbf{Weight} & \textbf{Run 1} & \textbf{Run 2 as reported} & \textbf{Run 2 recomputed}\\
\midrule
$\Snet$ & 0.30 & 0.948 (28.4) & 0.550 proxy (16.5) & 0.550 proxy (16.5)\\
$\Sbib$ & 0.25 & 0.864 (21.6) & 0.868 (21.7) & 0.868 (21.7)\\
$\Ssem$ & 0.20 & 0.608 (12.2) & 0.872 (17.4) & 0.872 (17.4)\\
$\Sproc$ & 0.25 & 0.790 (19.8) & 0.702 (17.6) & 0.539 (13.5)\\
\textbf{QualityScore (grade)} & & \textbf{81.9 (A)} & \textbf{73.2 (B)} & \textbf{69.1 (B)}\\
\midrule
\multicolumn{5}{@{}l}{\textit{Run 1 before the corpus-improvement cycle (79 papers)}}\\
$\Snet$ & 0.30 & 0.717 (21.5) & & \\
$\Sbib$ & 0.25 & 0.803 (20.1) & & \\
$\Ssem$ & 0.20 & 0.587 (11.7) & & \\
$\Sproc$ & 0.25 & 0.851 (21.3) & & \\
QualityScore (grade) & & 74.6 (B) & & \\
\midrule
\multicolumn{5}{@{}l}{\textit{Run 1 final score with the network computed differently (other sub-scores fixed)}}\\
Coupling threshold $\theta=2$ instead of 3 & & 74.0 (B) & & \\
Co-citation modularity instead of coupling & & 64.3--64.5 (C) & & \\
Louvain seed varied (100 seeds) & & 81.5--81.9 (A) & & \\
Interval for $\Snet \in [0,1]$ & & [53.5, 83.5] & & \\
\midrule
\multicolumn{5}{@{}l}{\textit{Run 2 under the missing-data rules}}\\
Coverage penalty, Eq.~\eqref{eq:penalty} with $c=0$ & & & 56.7 (C) & 52.6 (C)\\
Weights renormalized over three sub-scores & & & 81.0 (A) & 75.2 (B)\\
Interval for $\Snet \in [0,1]$ & & & [56.7, 86.7] & [52.6, 82.6]\\
\bottomrule
\end{tabularx}
\end{table}

\textit{Run 1.} Before improvement, the score of 74.6 (B) combined a strong process sub-score (inclusion rate 41.1\%, inside the band) with a weak semantic one (0.587): only 46\% of the 122 keyword-map terms occurred at least twice in the accepted abstracts, and the four outcome clusters overlapped (mean inter-cluster cosine similarity 0.593). The network sub-score rested on reference lists for 39 of the 79 papers, of which 22 entered the coupling graph. The quality report turned these values into specific actions. It proposed snowballing the nine heavily cited missing works, which it estimated would raise the h-index from 23 to about 30, and a query for 2018--2022 and non-LLM chatbots, whose expected score gain was small (about 0.3 points) but which addressed the corpus's inability to say anything about chatbots before ChatGPT. After the improvement cycle, the score rose by 7.3 points to 81.9 (A), the h-index to 35 (each citation count is the maximum of the value recorded at retrieval and the OpenAlex count, including, for one preprint, the count of its published version), the number of pre-2023 papers from 1 to 19 and the share of papers with reference lists to 58\%. Most of the gain came from $\Snet$ (+6.9 points), and part of it from fragmentation rather than better structure: the pre-LLM papers formed small coupling components of their own, which modularity rewards. $\Sproc$ lost 1.5 points because the broader search lowered the inclusion rate to 25.8\%, below the 30--60\% band, so two components of the index moved in opposite directions under the corrective action that the index itself had recommended. The grade also depends on the network computation: with the utility's default coupling threshold ($\theta=2$), the same corpus scores 74.0 (B), and a co-citation-based network gives 64.3 or 64.5 (C) in two executions of the notebook, because that Louvain call is unseeded, whereas 100 seeds of the coupling partition move the score by at most 0.4 points. The quality agent seeded the Louvain call during this run, a local change to \file{quality\_utils.py}. This is the intended diagnostic loop of checkpoint~B, in which a low component led to a specific corrective action and the score responded; it also shows that the grade alone would have hidden both the trade-off between components and the threshold dependence.

\textit{Run 2.} The reported score of 73.2 (B) rests on two departures from the specification. First, $\Snet$ is a proxy (0.55), not a coupling-based value. Second, the quality agent computed the components in its own code rather than with \file{quality\_utils.py}. It scored the inclusion-rate term as 0.638, whereas Eq.~\eqref{eq:phi} gives 0.153 for $r = 0.939$, and it counted keyword-map terms present in at least one abstract (26 of 31) instead of terms occurring at least twice. Recomputing $\Sproc$ with Eq.~\eqref{eq:phi} gives 0.539 and a composite of 69.1, still grade B. Under the coverage penalty of Eq.~\eqref{eq:penalty}, the same corpus would be graded C (52.6); with weights renormalized over the three available sub-scores, B (75.2). The network sub-scores of the two runs are therefore not comparable: Run~1's value comes from actual coupling data, Run~2's from a proxy. The other components of Run~2 are high: an h-index of 29, low inter-cluster similarity (0.178) and low redundancy (0.043). The semantic values, however, were computed on the one-sentence snippets that replaced the abstracts (Section~\ref{sec:setup:run2}); such short texts share few terms, which probably lowers both similarities and raises $\Ssem$.

\textit{Weight sensitivity.} To gauge how much the grades depend on the uncalibrated weights, we perturbed them around their nominal values (Appendix~\ref{app:qs}). Shifting any single weight by $\pm 0.05$, with the others rescaled to keep their sum at one, moves Run~1 (after improvement) between 80.6 and 83.3, always grade A, and Run~2 between 67.9 and 70.3, always grade B. With weights drawn from a Dirichlet distribution centered on the nominal weights (standard deviation of about 0.04 per weight), Run~1 keeps grade A in 94\% of the draws and Run~2 grade B in all of them; with a standard deviation of about 0.09, these shares fall to 77\% and 88\%. Equal weights give 80.2 (A) and 70.7 (B). The weights therefore matter mainly for corpora close to a grade threshold, as Run~1 is (1.9 points above the A threshold), whereas the coupling threshold and the treatment of missing network data (lower blocks of Table~\ref{tab:qsresults}) move the score more than any of these perturbations.

\subsection{Manuscript generation and internal review}\label{sec:results:review}

Run~1 produced a manuscript of about 9,500 words in Sections 1--13 (excluding citation keys and figure captions). It cites all 100 included studies through 461 citations, embeds six figures and adds a supplementary evidence-weighting table and an appendix with the search strings. The reviewer agent ran its checklist with scripts. All citation keys were present in the bibliography and the dataset, the PRISMA counts matched the curation reports and the flow diagram, and the 39 verbatim quotations matched their sources; it also spot-checked 68 numbers and about 20 study-level claims. It found no critical issue and 14 minor ones, among them a factual error (a section stated that all but one of its studies were LLM-based, whereas the data showed 22 LLM-based, one pre-LLM and one mixed-tool study), two overstated generalizations, a study design labelled more strongly than its source supports, two wrong figure references and six year discrepancies between the bibliography and the dataset. Its verdict was minor revision, and one revision round addressed the 14 issues. The export agent then produced a Word file (six embedded figures, 100 references), a LaTeX source and an 11-slide deck that passed validation, and the conductor converted the Word file to a 44-page PDF. Two defects surfaced only after the review. The titles drawn into the five generated figures were numbered 1--5, whereas the manuscript numbered them 2--6 because the PRISMA diagram had become Figure~1; the conductor noticed the mismatch in the PDF and regenerated the figures without numbers. The LaTeX source did not compile until three problems introduced by the BibTeX and export agents were corrected (Table~\ref{tab:failures}). Run~2's manuscript (about 8,650 words, five embedded figures, 28 cited works) was exported to Word and a 14-slide deck and received the same verdict. The reviewer asked for two additions, a statement that the review was not prospectively registered and a note on publication bias, and listed two factual claims to verify before submission. In its own PRISMA check, it also accepted QualityScore as a proxy for the assessment of risk of bias, which QualityScore is not designed to measure.

These verdicts should not be read as evidence of manuscript quality. The reviewer is an LLM of the same family as the author agents, and LLM evaluators are known to favor their own generations \citep{panickssery2024llm,zheng2023judging}. The most useful part of the review step is its checklist of verifiable properties, such as citations present in the dataset and counts consistent with the decision log. These checks caught the factual count error of Run~1, but neither its figure--caption drift nor the count discrepancies of Run~2, which surfaced only when the conductor inspected the exported PDF or when we compared the artifacts (Section~\ref{sec:results:failures}).

\subsection{Evidence on AI chatbots in higher education (Run 1 output)}\label{sec:results:chatbots}

\begin{figure}[pos=tbp]
\centering
\includegraphics[width=0.95\linewidth]{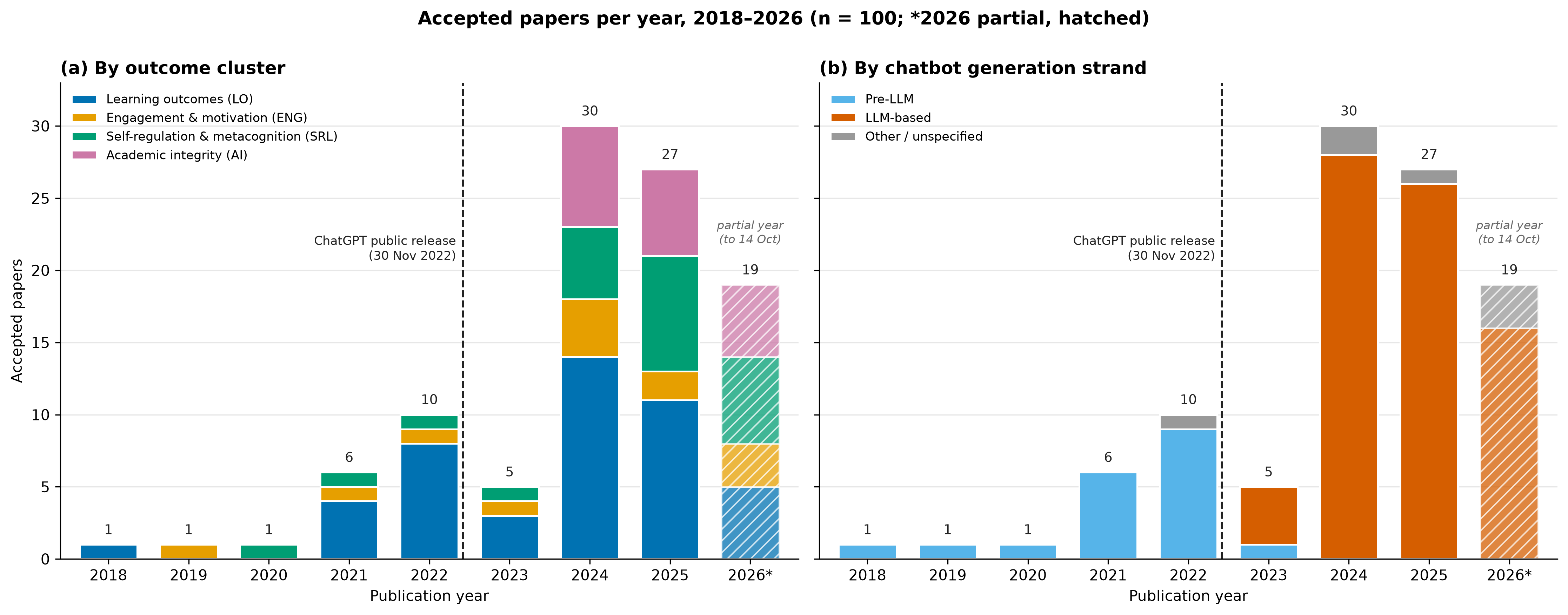}
\caption{Run~1: included studies per publication year ($n=100$), (a) by outcome cluster and (b) by chatbot generation strand, as generated by the figure agent. The dashed line marks the public release of ChatGPT (30~November~2022); the 2026 bar is hatched because the run took place in October 2026. Before the improvement cycle, the corpus held one study published before 2023; the second retrieval round and backward snowballing added 18.}\label{fig:run1timeline}
\end{figure}

\begin{figure}[pos=tbp]
\centering
\includegraphics[width=0.75\linewidth]{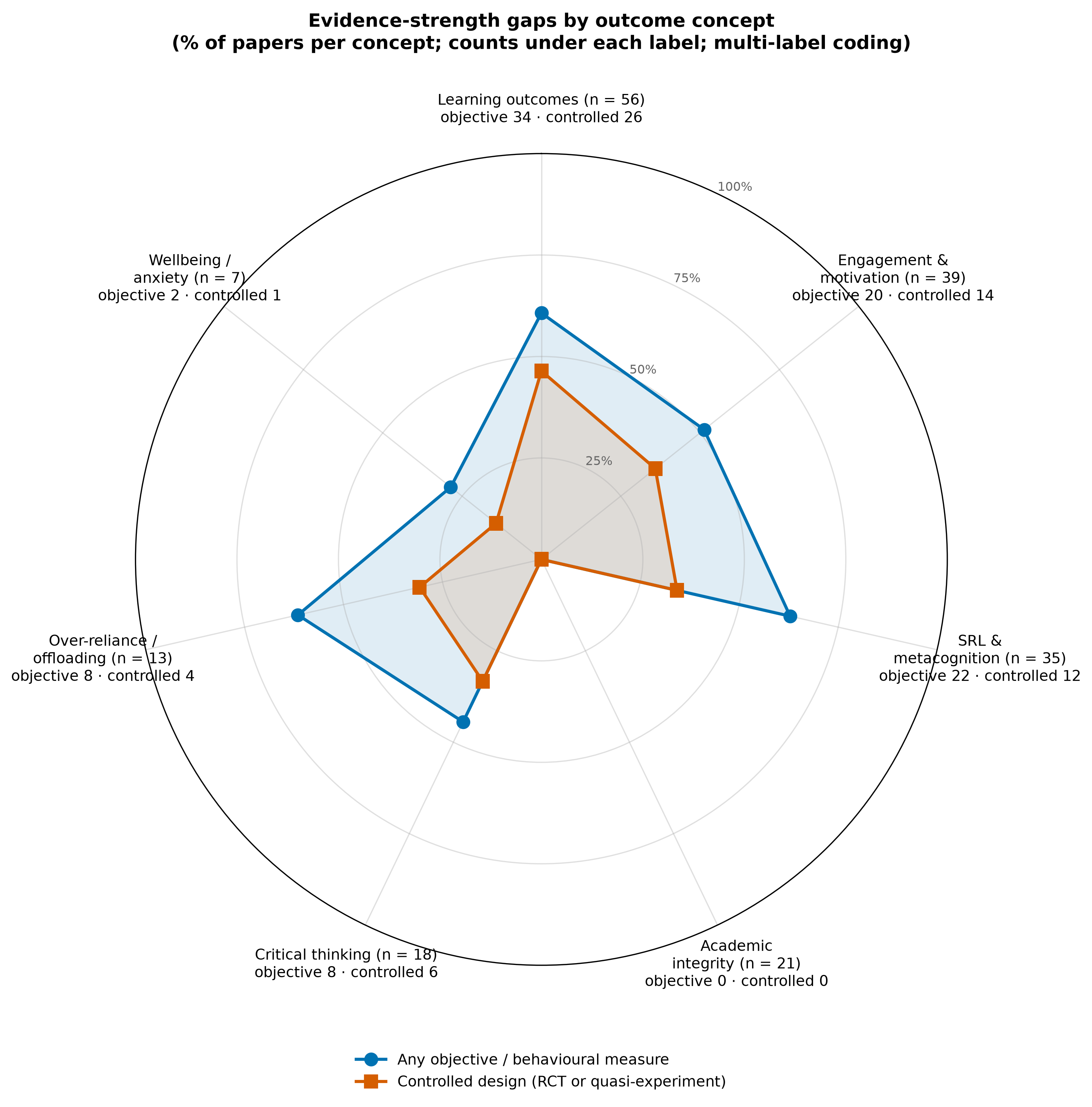}
\caption{Run~1: evidence strength by outcome concept, as generated by the figure agent from the gap report. Each axis gives the share of studies coded with the concept (multi-label coding; counts under each label) that use an objective or behavioral measure and that use a controlled design (randomized or quasi-experimental). No academic-integrity study uses either.}\label{fig:run1radar}
\end{figure}

Figures~\ref{fig:run1timeline} and \ref{fig:run1radar} reproduce two of the five figures generated during Run~1, and Table~\ref{tab:run1findings} lists the main findings of its review, checked against the run's dataset, gap report and limits report. The corpus shows a generational break (Fig.~\ref{fig:run1timeline}). The 19 pre-LLM studies, 18 of them published in 2018--2022, test rule-based, retrieval-based, neural or unspecified chatbots, whereas the 74 LLM-based studies all date from 2023 onwards and mostly concern ChatGPT. Learning outcomes are the most studied and best measured outcome, and academic integrity the least (Fig.~\ref{fig:run1radar}). Without the improvement cycle of checkpoint~B, the review could not have covered the first of these two generations: the first-round corpus contained a single study published before 2023.

\begin{table}[pos=tbp]
\caption{Main findings of the Run~1 review of AI chatbots and university students' learning, as produced by PARIS and checked against the run's dataset, gap report and limits report.}\label{tab:run1findings}
\footnotesize
\begin{tabularx}{\textwidth}{@{}P{3.6cm}Y@{}}
\toprule
\textbf{Finding} & \textbf{Evidence}\\
\midrule
Two chatbot generations, never compared & 19 pre-LLM studies (2018--2023) and 74 LLM-based studies (2023--2026); no study compares a rule-based chatbot with an LLM, and the pre-LLM strand ends in 2023, possibly in part because of the queries (Fig.~\ref{fig:run1timeline})\\
Learning effects are measured over short horizons & 26 controlled studies of learning outcomes, but 7 of the 11 randomized trials are single-session tasks; none of the 56 full texts reports a delayed retention test, and 1 of 48 exposure studies lasts longer than a semester\\
LLM outcomes are mostly self-reported & In the LLM strand, 42 studies rely on self-report only and 26 include an objective measure; experimental studies have a median of 61 participants, and one study was preregistered\\
Concern about over-reliance lacks causal evidence & 13 studies of over-reliance and cognitive offloading, all on LLMs, with four short controlled tests and no validated measure\\
Academic integrity is studied through attitudes only & 21 studies coded for integrity (2024--2026), 13 of them cross-sectional surveys, with no controlled design, objective measure or intervention test (Fig.~\ref{fig:run1radar})\\
Maturity and reproducibility are low & Mean evidence-maturity level 4.04 on the limits agent's adapted 1--9 scale (91 rateable studies; pre-LLM 5.18, LLM 3.79); open code in 4 and open data in 7 of the 100 studies; prompts or code reported in 4 of 23 full-text LLM interventions\\
\bottomrule
\end{tabularx}
\end{table}

\subsection{The landscape of AI-assisted systematic review methods (Run 2 output)}\label{sec:results:landscape}

\begin{figure}[pos=tbp]
\centering
\includegraphics[width=0.95\linewidth]{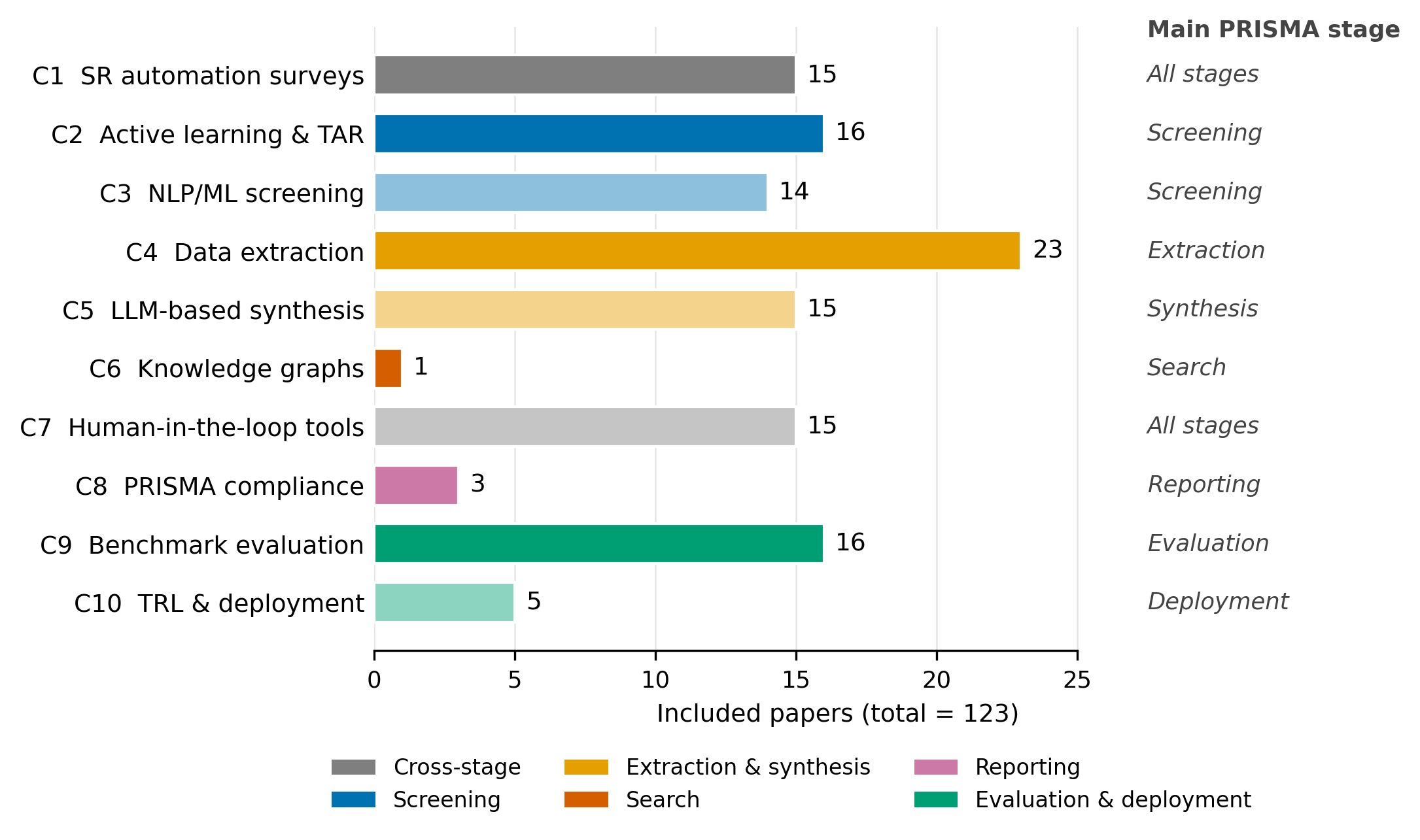}
\caption{Run~2: thematic clusters of the corpus on AI-assisted SR methods produced by PARIS, with the number of included papers per cluster in the final dataset (123 papers) and the review stage each cluster mainly addresses. Colors group clusters by stage, as in Fig.~\ref{fig:network}. Redrawn from the run's dataset; the figure generated during the run used the curation report's cluster table, which sums to 117.}\label{fig:taxonomy}
\end{figure}

\begin{figure}[pos=tbp]
\centering
\includegraphics[width=0.95\linewidth]{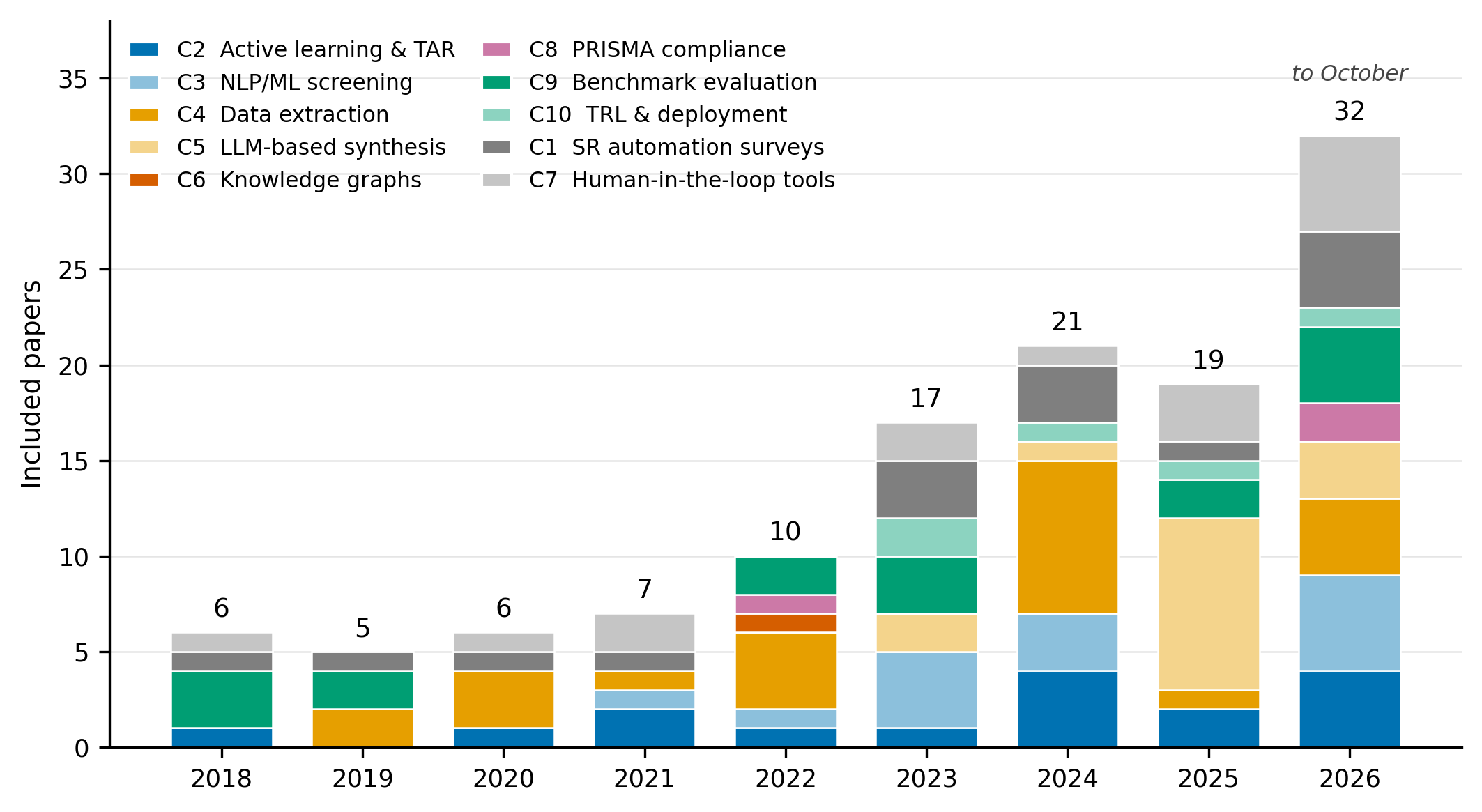}
\caption{Run~2: included papers per year and cluster (stacked bars; yearly totals printed above the bars). The 2026 count covers January to October, when the run took place.}\label{fig:timeline}
\end{figure}

\begin{figure}[pos=tbp]
\centering
\includegraphics[width=0.95\linewidth]{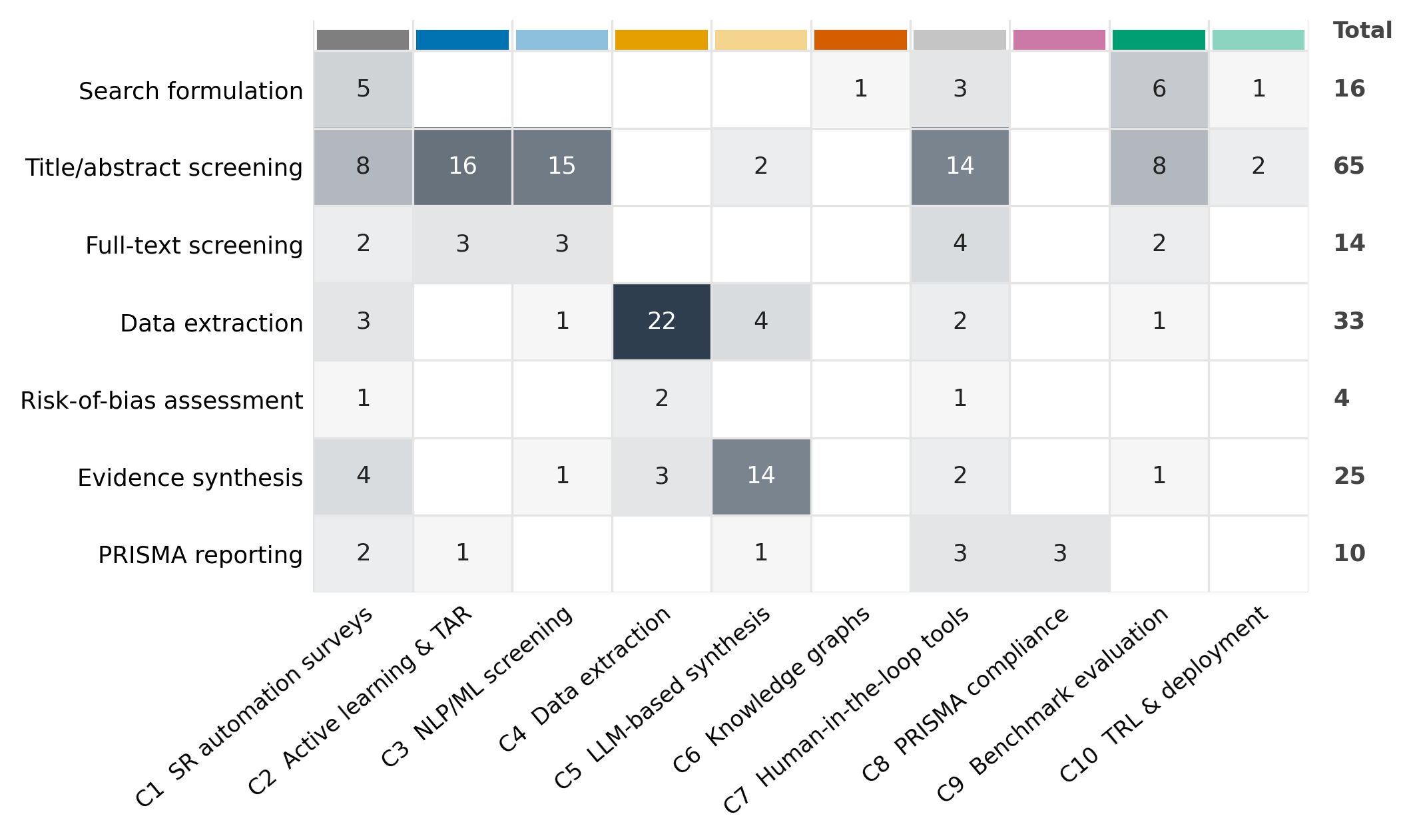}
\caption{Run~2: number of papers addressing each review stage (rows) within each thematic cluster (columns), as coded by the figure agent. Row totals are printed on the right; a paper may address several stages, so they do not sum to the corpus size, and the column sums are not cluster sizes. The colored strip identifies the clusters as in the other figures. The stage coding used the seven stages shown and is not stored in the dataset; the counts are transcribed from the figure produced during the run.}\label{fig:heatmap}
\end{figure}

\begin{figure}[pos=tbp]
\centering
\includegraphics[width=0.9\linewidth]{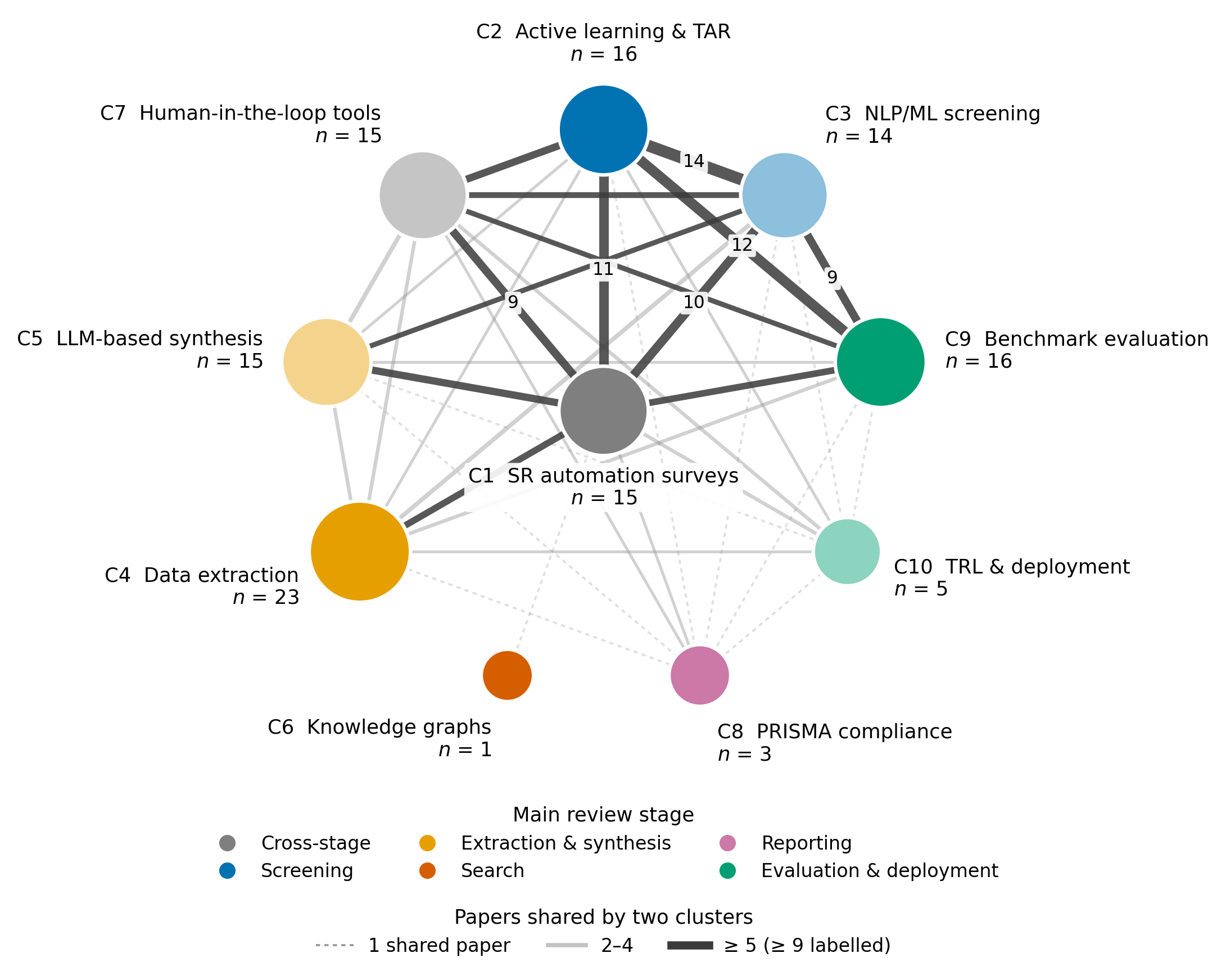}
\caption{Run~2: cluster co-occurrence network, redrawn from the gap agent's intersection matrix. Node area increases with the number of included papers per cluster ($n$) and colors give the main review stage, as in Fig.~\ref{fig:taxonomy}. Edges join cluster pairs estimated to share at least one paper, with width proportional to that number (values of 9 or more are printed). These counts are estimates made by the gap agent from abstracts and cluster assignments, not computed overlaps. The survey cluster (C1), linked to all others, is placed at the center; the knowledge-graph cluster (one paper) shares a single paper with C1 and none with the others.}\label{fig:network}
\end{figure}

Figures~\ref{fig:taxonomy}--\ref{fig:network} show four views of the Run~2 corpus, redrawn in a common style from the run's dataset, gap report and figure-agent output, with one color per cluster throughout, and Table~\ref{tab:findings} lists its main findings, checked against the run's dataset and reports. The corpus grew from 5--7 papers per year in 2018--2021 to 32 papers in 2026, a year still in progress at the time of the run (Fig.~\ref{fig:timeline}). Screening dominates: 65 of 123 papers address title and abstract screening in the stage coding of Fig.~\ref{fig:heatmap}. LLM-based synthesis (C5) appears only from 2023 and grows fastest. Risk-of-bias assessment and PRISMA reporting are the least covered stages.

\begin{table}[pos=tbp]
\caption{Main findings of the Run~2 review of AI-assisted SR methods, as produced by PARIS and checked against the run's dataset and reports.}\label{tab:findings}
\footnotesize
\begin{tabularx}{\textwidth}{@{}P{3.6cm}Y@{}}
\toprule
\textbf{Finding} & \textbf{Evidence}\\
\midrule
Screening dominates & 65/123 papers address title/abstract screening in the stage coding of Fig.~\ref{fig:heatmap}; the two screening clusters (C2, C3) hold 30 papers in the dataset\\
LLM synthesis grows fastest & C5 (15 papers in the dataset) has none before 2023, then 2, 1, 9 and 3 papers in 2023--2026 (Fig.~\ref{fig:timeline})\\
Appraisal and reporting are under-covered & Risk of bias: 4 papers in the stage coding, 2 of them dedicated (an evaluation of RobotReviewer and a GPT-4 study of ROBINS-I); meta-analysis: 2 dedicated papers; PRISMA reporting: 10 papers, 3 dedicated (C8)\\
Deduplication and protocol registration are absent from the corpus & No included paper addresses either, although automated deduplication tools exist \citep{borissov2022deduklick}; this points to a recall limit of the run's queries rather than a gap in the literature\\
Most work is below deployment maturity & In the limits agent's heuristic TRL matrix (117 papers), 14 papers reach TRL 7--9, spread over eight clusters, 4 of them review tools (C7)\\
Knowledge-graph approaches are marginal & One paper (C6), which shares a single paper with the survey cluster (Fig.~\ref{fig:network})\\
Reproducibility is uneven & Among the 29 included papers with full text, regular-expression detection found code links in 19 (66\%) and data statements in 15 (52\%); the limits agent's abstract-based estimates for the whole corpus are 34\% and 50\%\\
\bottomrule
\end{tabularx}
\end{table}

\subsection{Support for PRISMA 2020 reporting}\label{sec:results:prisma}

Table~\ref{tab:prisma} maps each PRISMA 2020 item to the PARIS artifacts that support it, based on the agent specifications and utilities of release 0.6. PARIS fully supports 8 items, partially supports 11 and does not support 8. Support is strongest for transparency of the search and selection (items 5--7 and 16), where verbatim query logs and per-record decision rationales exceed what many manual reviews report, provided the agents follow their specifications (Section~\ref{sec:results:failures}). It is weakest for appraisal and certainty (items 11, 14, 15, 18, 21 and 22), which PARIS does not implement. In Run~1, the flow diagram separated database searching from citation searching and reported the post hoc exclusions, as PRISMA 2020 asks of reviews that use several identification methods, but it omitted the counts of reports sought and not retrieved. Because PARIS syntheses are thematic maps of a field rather than effect estimates, its outputs are closer to scoping reviews or systematic mapping studies, for which PRISMA-ScR is the more appropriate reporting guideline \citep{tricco2018prismascr}, unless appraisal modules are added.

\begin{table}[pos=tbp]
\caption{Support for the 27 PRISMA 2020 items by PARIS release 0.6 (F: full; P: partial; --: not supported), assessed by the authors from the agent specifications.}\label{tab:prisma}
\footnotesize
\begin{tabularx}{\textwidth}{@{}P{0.45cm}P{3.6cm}P{0.5cm}Y@{}}
\toprule
\textbf{\#} & \textbf{Item} & & \textbf{PARIS artifact or gap}\\
\midrule
1 & Title & P & Written by the synthesizer; identification as a systematic review not enforced\\
2 & Abstract & P & At most 250 words with corpus size and QualityScore; PRISMA abstract checklist not enforced\\
3 & Rationale & F & Introduction derived from the research goal and concept map\\
4 & Objectives & F & Explicit research questions\\
5 & Eligibility criteria & F & Scope boundaries (scoper) and criteria applied (curator)\\
6 & Information sources & F & Sources, API status and extraction date (strategist)\\
7 & Search strategy & F & Every query URL logged verbatim with counts\\
8 & Selection process & P & One LLM screener; human decisions only for borderline records\\
9 & Data collection process & P & Heuristic full-text extraction and LLM summaries; no human verification step\\
10 & Data items & P & Fixed dataset schema; outcomes per question not specified\\
11 & Study risk of bias assessment & -- & Not implemented (TRL and reproducibility maps are not risk-of-bias tools)\\
12 & Effect measures & -- & No quantitative synthesis\\
13 & Synthesis methods & P & Thematic clustering and narrative synthesis described; no statistical methods\\
14 & Reporting bias assessment & -- & Not implemented\\
15 & Certainty assessment & -- & Not implemented\\
16 & Study selection & F & Flow diagram from the decision log; rationale for every excluded record\\
17 & Study characteristics & F & Master dataset, cluster tables and figures\\
18 & Risk of bias in studies & -- & Not implemented\\
19 & Results of individual studies & P & Metadata and extracted sentences per study; no effect estimates\\
20 & Results of syntheses & P & Narrative syntheses per cluster\\
21 & Reporting biases & -- & Not implemented\\
22 & Certainty of evidence & -- & Not implemented\\
23 & Discussion & F & Interpretation; limitations of evidence (limits agent) and of the process; implications\\
24 & Registration and protocol & P & Time-stamped run template and scoping outputs form a protocol-like record; no registration\\
25 & Support & -- & Supplied by the authors\\
26 & Competing interests & P & Template statement to be completed by the authors\\
27 & Availability of data, code and materials & P & All artifacts persisted and shareable; statement written by the authors\\
\bottomrule
\end{tabularx}
\end{table}

\subsection{Observed failure modes}\label{sec:results:failures}

Table~\ref{tab:failures} lists the failures observed in the two runs, found by auditing their artifacts, with their root causes and the mitigations already implemented or proposed. Three patterns stand out. First, rules that exist only as instructions to an LLM can be skipped silently or improvised. In Run~2, the strategist did not paginate and stored snippets instead of abstracts, the analyst produced no notebook, PDF retrieval ran before curation on arXiv records only, the quality agent bypassed the shared utilities, and no checkpoint decision was recorded. In Run~1, a template parameter capped the first retrieval round, the conductor had to improvise the corpus-improvement path of checkpoint~B, and checkpoint~A was delegated to the conductor itself. Second, every problem that releases 0.5 and 0.6 fixed during development was fixed by moving a rule from prose into an explicit procedure: the deduplication key order, the contamination guard, count re-derivation and figure embedding. Where Run~1's conductor re-derived the counts from the decision logs at every step, as release 0.5 specifies, no count drifted between the curation artifacts. Third, numbers written by LLM agents remain the least reliable artifacts of a run. The Run~2 curator's summary reported 114 accepted papers, its cluster table 117 and its decision log 125, while the dataset holds 123, and none of these discrepancies was caught by the reviewer agent. In Run~1, agents applied different definitions of the same quantity (13 or 19 pre-LLM studies), and the reviewer caught one wrong count in the manuscript.

\begingroup
\footnotesize
\setlength{\LTcapwidth}{\textwidth}
\begin{xltabular}{\textwidth}{@{}P{3.0cm}YYP{3.6cm}@{}}
\caption{Failure modes observed in the case studies, with root causes and mitigations (implemented in the stated release, or proposed).}\label{tab:failures}\\
\toprule
\textbf{Failure (run)} & \textbf{Observed effect} & \textbf{Root cause} & \textbf{Mitigation}\\
\midrule
\endfirsthead
\multicolumn{4}{@{}l}{\textit{\tablename~\thetable{} (continued)}}\\
\toprule
\textbf{Failure (run)} & \textbf{Observed effect} & \textbf{Root cause} & \textbf{Mitigation}\\
\midrule
\endhead
\midrule
\multicolumn{4}{r@{}}{\textit{Continued on next page}}\\
\endfoot
\bottomrule
\endlastfoot
Incomplete retrieval (2) & Five or six records captured per OpenAlex query; off-topic records dropped without a log & Agent fetched result pages with a web tool instead of paginating the API & Proposed: deterministic, paginated retrieval client that logs every record\\
Retrieval capped by the run template (1) & First round limited to the first relevance-ranked page of each query; reported totals logged & The template's ``about 200 papers'' parameter conflicts with the strategist's full-pagination rule & Proposed: reconcile template and specification; report truncation as a recall limit\\
Abstracts replaced by snippets (2) & Abstract field holds 43--109-character paraphrases; screening, $\Ssem$, keyword coverage and abstract-based TRL rest on them & Truncated web-fetch output summarized by the agent & Proposed: store API abstracts verbatim and check their length\\
Analysis notebook not produced (2) & No executed notebook in the run folder; analyses recomputed ad hoc by later agents & Conductor marked the step done with the notebook still pending & Proposed: block step completion until the executed notebook exists\\
Unrecorded checkpoint decision (2) & Tracker reports outline approval; outline file still awaiting approval & Approval recorded as free text by the agent & Proposed: hard gate requiring a human-written approval file\\
Delegated checkpoint (1) & Both rounds of checkpoint A (72 borderline records, 6 rechecks) decided by the conductor under a rule it wrote; 15 of the 72 decisions departed from the curator's recommendation & Checkpoints are enforced by instruction; nothing distinguishes a delegated decision from a human one & Proposed: record who decided; human-written approval file\\
Count drift (2) & 114, 117, 125 and 123 included papers in four artifacts; two accepted records lost & Counts written by agents rather than recomputed & Re-derivation from decision logs (0.5); proposed: consistency checks in code\\
Stale source of truth after a delta cycle (1) & Notebook quality section read the first-round curation report (79 included) instead of the combined report (100) and failed & Append-only delta files leave no single file that holds the current corpus & Proposed: regenerate a consolidated corpus manifest after each delta\\
Improvised improvement path (1) & Snowballing and date-bounded queries run through conductor-written prompts & No agent implements the improvement path of checkpoint B; the updater only fetches newer publications & Proposed: an improvement agent with its own specification\\
Specification drift in scoring (2) & Inclusion-rate term 0.638 instead of 0.153; coverage counted at one occurrence instead of two & Quality agent did not use \file{quality\_utils.py} & Proposed: scores computed only by the shared utilities\\
Silent parsing fallback (1) & Keyword coverage would have defaulted to 0.5 without warning & Scoper table format differs from the regular expression in \file{quality\_utils.py} & Proposed: validate the parse and fail loudly\\
Definition drift between agents (1) & Pre-LLM strand counted as 13 (quality agent) or 19 (gap and limits agents) studies & Derived codes are not stored in the dataset & Proposed: store derived codes as dataset columns\\
API outage (2) & $\Snet$ not computable; ad hoc proxy & Single source of reference lists; no fixed missing-data rule & Coverage penalty (0.5); proposed: interval reporting\\
Missing credentials (1, 2) & Semantic Scholar HTTP 429 in both runs; in Run 1 no Elsevier, IEEE or Unpaywall access and full text for 30.6\% of second-round records & Keys unavailable; contact address declined & Key pre-flight (0.5); proposed: report the coverage lost per missing source\\
Out-of-order full-text retrieval (2) & PDFs fetched before curation, arXiv only, capped at 30 & Conductor did not follow the step order and source cascade & Proposed: enforce order and cascade in code\\
Partial write on interruption (1) & An interrupted translator run had rewritten the extraction list without writing its report & Agents edit input artifacts in place; persistence gates check outputs only & Proposed: write annotations to new files; check input hashes\\
Figure--caption drift (1, 2) & Run 2: captions describing other plots. Run 1: titles drawn into the images numbered 1--5, manuscript numbers 2--6 & Figures produced ad hoc or before the manuscript's figure order is known; titles burned into images & Proposed: captions and numbers generated with the figure code at export\\
Figures not embedded (2) & Exports without figures & Text-only figure references & Image syntax and figure embedder (0.6)\\
Domain-specific defaults (1) & Figure utilities hard-wired to another domain's clusters, platforms and keywords, one saving at 150 dpi; Run 1's figure agent wrote its own code & Utilities extracted from a development-run notebook & Proposed: parameters from the scoper output\\
Personal contact address in shared code (1) & OpenAlex utility sends a developer's e-mail address by default; Run 1's agents bypassed it & Development default left in a shared utility & Proposed: read the address from the run configuration, none by default\\
Non-compiling LaTeX export (1) & BibTeX parse errors and undefined table commands at first compilation & Comment lines containing \texttt{@} in the bibliography; custom preamble missing pandoc definitions; no compilation check & Proposed: compile the LaTeX output in the export step\\
\end{xltabular}
\endgroup
\section{Discussion}\label{sec:discussion}

\subsection{Answers to the research questions}\label{sec:discussion:rq}

\textbf{RQ1.} A fixed workflow of LLM agents that communicate only through persisted files makes intermediate decisions auditable after the fact: queries, duplicates, scores and rationales, quota decisions, metrics, outlines and review checklists all remain in the run folder. The audit of Run~2 shows both the value and the limit of this design. The artifacts made every deviation detectable, but the persistence checks did not prevent them, because they test that files exist rather than that agents followed their specifications. Human authority at the three checkpoints is likewise enforced by instructions rather than by mechanism. Run~1 adds a second limit: a checkpoint protects human authority only if the researcher exercises it. Its researcher delegated both rounds of checkpoint~A to the conductor, which wrote its own inclusion rule, decided 72 borderline records and six rechecks, and departed from the curator in 15 of the 72 borderline decisions. The artifacts record the delegation and every reason, but the pipeline does not distinguish a delegated decision from a human one.

\textbf{RQ2.} QualityScore provided an actionable signal in Run~1. The quality report's diagnosis (one pre-2023 paper, no non-LLM chatbot, nine missing highly cited works) led to backward snowballing and a targeted query, a 7.3-point gain and a corpus that covers both chatbot generations. The same cycle showed that components can move in opposite directions: the broader search raised $\Snet$ and $\Sbib$ but lowered $\Sproc$. The grade, however, is sensitive to weights that have not been calibrated, mainly for corpora near a grade threshold such as Run~1, to the coupling threshold (81.9 or 74.0 for the same Run~1 corpus) and even more to the handling of missing data: the same Run~2 corpus scores between 52.6 and 82.6 depending on the network sub-score, and 69.1 or 73.2 depending on whether the specified formula is applied. Its components are more informative than its letter grade, and they should be reported together.

\textbf{RQ3.} Both runs completed the workflow from a one-sentence goal to exported documents in two unrelated domains. Run~1 spanned 14--26~October, with long idle periods between working sessions; its wall-clock time and token use were not logged. Completing the workflow is not the same as executing it as specified. The failure modes in Table~\ref{tab:failures} show where LLM instructions are insufficient and deterministic procedures are needed, starting with retrieval and counting.

\textbf{RQ4.} PARIS supports the search and selection items of PRISMA 2020 strongly and the appraisal and certainty items not at all. Its distinctive features relative to existing tools are corpus-level quality gating, artifact-level provenance and coverage of domain-agnostic scoping and mapping reviews. Several commercial platforms cover more conventional stages, clinical agentic systems offer much stronger validation, and recent pipelines also generate manuscripts end to end.

\subsection{Implications for practice}\label{sec:discussion:practice}

Three recommendations follow for teams considering an agentic review pipeline. First, use it where its strengths lie. PARIS is best suited to scoping, mapping and first drafts in fast-moving fields. It can be combined with specialized tools for the decisions it handles least rigorously: screening with principled stopping rules \citep{vandeschoot2021asreview,callaghan2020statistical} and risk-of-bias assessment \citep{marshall2016robotreviewer,lai2024assessing}. Second, report its outputs as what they are, typically scoping reviews or systematic maps, following PRISMA-ScR \citep{tricco2018prismascr}. Third, audit the artifacts, not the summaries. Every number in an LLM-written report should be recomputed from the dataset, and the query log should be checked for complete retrieval before the corpus is trusted.

\subsection{Responsible use, authorship and research integrity}\label{sec:discussion:ethics}

AI tools cannot be authors because they cannot take responsibility for a work, and publishers require that their use in writing be declared and their use in the research process be described in the methods \citep{elsevier2026genai}. The joint position statement of Cochrane, the Campbell Collaboration, JBI and the Collaboration for Environmental Evidence asks for transparency, human oversight and evidence that AI tools perform adequately before they are used in evidence synthesis \citep{flemyng2025position}. ARISMA similarly requires every consequential decision to remain human-interpretable, human-auditable and human-accountable \citep{tourchimoghaddam2026arisma}. PARIS's artifact trail supports transparency, and its checkpoints provide a minimum of human oversight. Its manuscripts are drafts: they require expert reading, verification of every claim and citation against the sources, and revision. Checkpoints also invite automation bias \citep{goddard2012automation}, because reviewers see the agent's recommendation before deciding; in Run~2, the researcher nonetheless overrode three of the curator's seven borderline recommendations. In Run~1, the researcher delegated checkpoint~A entirely, so that a second LLM, the conductor, took the reviewer's place for the borderline records; we report this as a use of the system that its design does not prevent. A system that lowers the cost of producing review-shaped documents could be misused to mass-produce low-quality reviews; we regard explicit disclosure, artifact sharing and journal policies requiring them as the main safeguards.

\subsection{Threats to validity}\label{sec:discussion:validity}

We follow the classification used in software engineering case-study research \citep{runeson2009guidelines}.
\textit{Construct validity.} QualityScore operationalizes adequacy through structural proxies whose weights are uncalibrated, and it does not measure recall. The internal reviewer verdict is not a measure of manuscript quality.
\textit{Internal validity.} Both runs were conducted by members of the author team, who also made, or in Run~1 delegated, the checkpoint decisions. The second run's research goal asked for a positioning of PARIS, which may have biased its scoping and synthesis. Run~1's conductor added instructions to the agent prompts that the specifications do not contain (Section~\ref{sec:setup:protocol}), and the Louvain call was patched during its quality step. Run~2's retrieval was incomplete and suffered an API outage. Run~1 is described here from its run folder by the same model that conducted it.
\textit{External validity.} Two runs in two domains, educational technology and SR methods, with two models from a single provider and two releases, do not support claims of generality, in particular for clinical reviews, non-English literatures or paywalled fields with low open-access coverage; without Unpaywall or publisher access, Run~1's second round obtained full text for only 30.6\% of its records.
\textit{Reliability.} Each topic was run once. LLM outputs vary between runs, and the Louvain step is unseeded in release 0.6, so repeated runs may differ.
\textit{Security.} Agents read untrusted abstracts and full texts, which exposes them to indirect prompt injection \citep{greshake2023indirect}; PARIS has no specific defense.

\subsection{Limitations}\label{sec:discussion:limitations}

Beyond the threats above, PARIS depends on a single LLM provider and on Claude Code, and is not portable to other providers without modification. Its network sub-score depends on OpenAlex. It implements neither risk-of-bias nor certainty assessment, nor meta-analysis. Several algorithms (retrieval and pagination, deduplication, quotas and count checks) exist only as prose specifications executed by LLM agents. The figure utilities carry domain-specific defaults from the development run from which they were extracted, and the shared OpenAlex utility sends a developer's contact address by default. Run cost and token consumption were not logged, so, unlike LUMEN \citep{huang2026lumen}, we cannot report the cost of a review.

\subsection{Future work}\label{sec:discussion:future}

The priorities follow from the analysis above:
(i) external validation against reference reviews. Replicating published reviews, as done for TrialMind, otto-SR and LUMEN \citep{wang2025trialmind,cao2025ottosr,huang2026lumen}, or evaluating PARIS on multi-stage benchmarks such as SciLitBench \citep{zabaleta2026scilitbench}, would measure the recall and precision of inclusion decisions and the accuracy of extracted data. Repeated runs would measure stability.
(ii) Deterministic implementations of retrieval, deduplication, quota and count checks, with every reported value computed from the dataset.
(iii) Mechanical enforcement of checkpoints through a run state machine that requires a human-written approval file before continuing.
(iv) Calibration of QualityScore against expert judgments, a sensitivity analysis extended to its saturation points, seeded or Leiden-based community detection \citep{traag2019louvain}, field-normalized citation indicators and a fixed missing-data rule.
(v) Risk-of-bias and certainty modules.
(vi) Provider abstraction with cost and token accounting.
(vii) PRISMA-ScR-aligned outputs.
(viii) Defenses against prompt injection.
(ix) A specified agent for the improvement path of checkpoint~B (backward snowballing and date-bounded delta queries), a consolidated corpus manifest after each delta cycle, and a compilation check of the LaTeX export.

\section{Conclusion}\label{sec:conclusion}

PARIS shows that a fixed workflow of LLM agents, coordinated exclusively through persisted and validated artifacts and interrupted by human checkpoints, can carry a review from a one-sentence research question to an exported manuscript draft in different domains while leaving a complete audit trail. QualityScore adds a pre-synthesis view of corpus adequacy whose components proved actionable, although its grade depends on uncalibrated weights and on how missing citation data are handled. The audits of both case studies show where the approach is fragile. Rules expressed only as instructions to an LLM were skipped or improvised, numbers written by agents drifted between artifacts, a human checkpoint was delegated back to the model, and only the persisted artifacts made these deviations visible. Moving such rules into deterministic procedures, and validating PARIS outputs against reviews conducted by independent human teams, are the steps needed before agentic review pipelines can be trusted beyond the drafting and mapping tasks for which PARIS is currently suited.

\section*{Data and code availability}
The PARIS source code (agent specifications, skills and Python utilities, release 0.6) is distributed under the MIT licence \citep{cruz2026paris}. The run artifacts of the two case studies (query logs, decision logs, datasets, notebooks, reports and generated manuscripts) are excluded from the code repository by design, and they will be made available later.

\section*{Acknowledgments}
The authors thank DAVI (Davi The Humanizers, Puteaux, France) for their support. The research project was supported by the Agence Nationale de la Recherche (ANR) under the Grant No ANR-24-LCV1-0008 (LAMAE).
This work also benefited from the support of the Dispositif Itinéraire Chercheur-Entrepreneur (ICE), funded by the Région Bourgogne-Franche-Comté.

\appendix

\section{Excerpt of an agent specification}\label{app:soul}

The following condensed excerpt of the curator specification (\file{curator\_agent/SOUL.md}) illustrates the declarative style; wording is shortened, and the scoring table and the output templates are abbreviated.

\begin{footnotesize}
\begin{verbatim}
# SOUL.md -- Curator Agent
## Mission
Apply structured inclusion/exclusion criteria to the raw extraction list. Score every
paper, enforce thematic balance, and surface borderline cases for human review.
## Phase 0 -- Pre-deduplication
Two papers are the same if they share: (1) arXiv ID, (2) DOI, (3) normalized title
(first 100 chars). Keep the most complete record; mark others duplicate_of: <kept_id>.
## Scoring Scale
0 out of scope | 1-3 tangential | 4-6 relevant | 7-8 highly relevant | 9-10 essential
Default acceptance threshold: score >= 4 (overridable in RUN.md).
## Cluster Balance
Flag any cluster with > 30% of accepted papers.
Soft cap: max_per_cluster = ceil(total_accepted / n_concepts x 1.5)
## Borderline Review Protocol
Score 3, or score 4 when uncertain -> borderline_review.md -> stop and notify the
Conductor; do not write curation_report.md before decisions are received.
## Completion Signal (two-phase)
Phase 1: CURATOR CHECKPOINT -- borderline_review.md written. N papers need human review.
Phase 2: CURATOR DONE -- curation_report.md written. N accepted, N rejected, ...
## Vibe
A rigorous peer reviewer with editorial judgment. [...] Borderline cases go to the
human -- you don't decide alone.
\end{verbatim}
\end{footnotesize}

\section{Worked QualityScore example}\label{app:qs}

For Run~1 after the improvement cycle, Eq.~\eqref{eq:qs} gives
\[
\QS = 100\,(0.30 \times 0.948 + 0.25 \times 0.86375 + 0.20 \times 0.6075 + 0.25 \times 0.7903) = 100\,(0.2844 + 0.2159 + 0.1215 + 0.1976) = 81.9,
\]
which is grade A. The components are $\Snet = \min(Q/0.7, 1)$, for a Louvain modularity $Q \approx 0.664$ on the 42 papers of the coupling graph ($\theta = 3$; 0.948 from the unrounded value); $\Sbib = \tfrac{1}{4}(1.000 + 0.782 + 0.673 + 1.000)$, where the h-index term ($h = 35$) and the foundational-rate term ($f = 0.29$) saturate; $\Ssem = \tfrac{1}{3}(0.403 + 0.521 + 0.898)$; and $\Sproc = \tfrac{1}{3}(0.859 + 0.727 + 0.785)$, where $\varphi(100/388) = 0.258/0.30 = 0.859$ and $\bar{m} = 6.28$. With $\theta = 2$, the modularity falls to 0.479 and $\Snet$ to 0.684, which gives 74.0 (B). Before the improvement cycle, the same formula gave $100\,(0.30 \times 0.717 + 0.25 \times 0.803 + 0.20 \times 0.587 + 0.25 \times 0.851) = 74.6$ (B).

For Run~2, the run's quality report gives the process components 0.638 (inclusion rate), 0.593 ($1-G_s$, with $G_s = 0.407$) and 0.873 ($\bar{m}/8$, with $\bar{m} = 6.98$), whose mean is the reported $\Sproc \approx 0.70$. With $r = 123/131 = 0.9389$, Eq.~\eqref{eq:phi} gives $\varphi(r) = 1 - (0.9389-0.60)/0.40 = 0.1527$, hence $\Sproc = (0.1527 + 0.5930 + 0.8725)/3 = 0.5394$. The non-network contributions then sum to $21.7 + 17.4 + 13.5 = 52.6$, so the composite ranges from 52.6 ($\Snet = 0$) to 82.6 ($\Snet = 1$); the proxy value 0.55 gives 69.1, and renormalizing the weights over the three available sub-scores gives $52.6/0.70 = 75.2$.

The weight sensitivity analysis of Section~\ref{sec:results:qs} uses the sub-scores of Table~\ref{tab:qsresults} (after-improvement values for Run~1, recomputed values for Run~2). The one-at-a-time analysis changes one weight $w_i$ by $\pm 0.05$ and rescales the other three proportionally so that the weights sum to one. The probabilistic analysis draws $10^5$ weight vectors from a Dirichlet distribution with parameters $\alpha w_i$, where $w_i$ are the nominal weights, for $\alpha = 100$ and $\alpha = 20$ (fixed seed), and reports the share of vectors that leave the grade unchanged.

\bibliographystyle{cas-model2-names}
\bibliography{paris}

\end{document}

%% file: figs/paris_architecture.tex
\begin{tikzpicture}[
  font=\sffamily\scriptsize,
  agent/.style={draw=black!70, rounded corners=2pt, fill=white, align=center,
                text width=21mm, minimum height=9mm, inner sep=1.5pt},
  cond/.style={agent, dashed},
  gate/.style={draw=orange!75!black, fill=orange!22, diamond, aspect=1.25,
               minimum size=8.5mm, inner sep=0pt, font=\sffamily\small\bfseries},
  gatelab/.style={font=\sffamily\tiny, align=center, text=orange!40!black},
  band/.style={rounded corners=4pt, draw=none},
  bandlab/.style={font=\sffamily\scriptsize\bfseries, anchor=north west, text=black!75},
  flow/.style={-{Stealth[length=1.8mm]}, semithick, black!75},
  loop/.style={-{Stealth[length=1.8mm]}, semithick, dashed, black!60},
  wide/.style={draw=black!60, rounded corners=3pt, align=left, inner sep=3pt},
  role/.style={font=\sffamily\tiny}
]
\def\xa{0}\def\xb{2.6}\def\xc{5.2}\def\xd{7.8}\def\xe{10.4}\def\xf{13.0}
\def\ya{0}\def\yb{-1.95}\def\yc{-3.9}\def\yd{-5.65}

\node[agent] (sco) at (\xa,\ya) {\textbf{1 Scoper}\\[-1pt]{\tiny concept map, PICO frame}};
\node[agent] (str) at (\xb,\ya) {\textbf{2 Strategist}\\[-1pt]{\tiny 6 APIs, logged queries}};
\node[cond]  (pre) at (\xc,\ya) {\textbf{2.5 PDF retriever}\\[-1pt]{\tiny open-access cascade}};
\node[cond]  (tra) at (\xd,\ya) {\textbf{3 Translator}\\[-1pt]{\tiny non-English abstracts}};
\node[agent] (cur) at (\xe,\ya) {\textbf{4 Curator}\\[-1pt]{\tiny dedup, 0--10 score, quotas}};
\node[gate]  (ga)  at (\xf,\ya) {A};
\node[gatelab, right=0.5mm of ga] {borderline\\papers};

\node[agent] (ana) at (\xf,\yb) {\textbf{5 Analyst}\\[-1pt]{\tiny CSV, notebook S1--S6}};
\node[agent] (bib) at (\xe,\yb) {\textbf{5b BibTeX}\\[-1pt]{\tiny Crossref / arXiv}};
\node[cond]  (pda) at (\xd,\yb) {\textbf{5.5 PDF analyzer}\\[-1pt]{\tiny full-text heuristics}};
\node[agent] (qua) at (\xc,\yb) {\textbf{6 Quality}\\[-1pt]{\tiny QualityScore (QS)}};
\node[agent] (gap) at (\xb,\yb) {\textbf{7 Gap finder}\\[-1pt]{\tiny five gap dimensions}};
\node[agent] (lim) at (\xa,\yb) {\textbf{7.5 Limits}\\[-1pt]{\tiny TRL, top-5 paths}};

\node[gate]  (gb)  at (\xa,\yc) {B};
\node[gatelab, below=0.4mm of gb] {QS + TRL};
\node[agent] (fig) at (\xb,\yc) {\textbf{8 Figures}\\[-1pt]{\tiny 5 figures, manifest}};
\node[agent] (out) at (\xc,\yc) {\textbf{9a Synthesizer}\\[-1pt]{\tiny outline}};
\node[gate]  (gc)  at (\xd,\yc) {C};
\node[gatelab, below=0.4mm of gc] {outline};
\node[agent] (dra) at (\xe,\yc) {\textbf{9b Synthesizer}\\[-1pt]{\tiny manuscript draft}};
\node[agent] (rev) at (\xf,\yc) {\textbf{10 Reviewer}\\[-1pt]{\tiny 10-dimension audit}};

\node[agent] (exp) at (\xf,\yd) {\textbf{11 Export}\\[-1pt]{\tiny DOCX, LaTeX, PPTX}};
\node[cond]  (emb) at (\xe,\yd) {\textbf{Figure embedder}\\[-1pt]{\tiny retrofit (optional)}};
\node[cond]  (upd) at (\xd,\yd) {\textbf{$\Delta$ Updater}\\[-1pt]{\tiny living-review deltas}};
\node[gate, minimum size=5mm, font=\sffamily\tiny\bfseries] (lg) at (\xa-0.55,\yd+0.12) {};
\node[anchor=west, font=\sffamily\tiny] at ([xshift=1mm]lg.east) {human checkpoint};
\draw[dashed, black!70, rounded corners=1pt] (\xa-0.8,\yd-0.38) rectangle (\xa-0.3,\yd-0.13);
\node[anchor=west, font=\sffamily\tiny] at (\xa-0.25,\yd-0.255) {conditional / optional step};
\draw[flow] (\xb+0.05,\yd+0.12) -- (\xb+0.55,\yd+0.12);
\node[anchor=west, font=\sffamily\tiny] at (\xb+0.6,\yd+0.12) {step order};
\draw[loop] (\xb+0.05,\yd-0.255) -- (\xb+0.55,\yd-0.255);
\node[anchor=west, font=\sffamily\tiny] at (\xb+0.6,\yd-0.255) {loop / retrofit};

\draw[flow] (sco) -- (str);  \draw[flow] (str) -- (pre);  \draw[flow] (pre) -- (tra);
\draw[flow] (tra) -- (cur);  \draw[flow] (cur) -- (ga);   \draw[flow] (ga)  -- (ana);
\draw[flow] (ana) -- (bib);  \draw[flow] (bib) -- (pda);  \draw[flow] (pda) -- (qua);
\draw[flow] (qua) -- (gap);  \draw[flow] (gap) -- (lim);  \draw[flow] (lim) -- (gb);
\draw[flow] (gb)  -- (fig);  \draw[flow] (fig) -- (out);  \draw[flow] (out) -- (gc);
\draw[flow] (gc)  -- (dra);  \draw[flow] (dra) -- (rev);
\draw[flow] (rev) -- node[right, font=\sffamily\tiny, align=left] {ACCEPT /\\MINOR} (exp);
\draw[loop] (rev.north) to[bend right=35] node[above, font=\sffamily\tiny] {revise ($\leq$3 rounds)} (dra.north);
\draw[loop] (emb) -- (exp);
\draw[loop] (gb.west) -- ++(-1.15,0) |- node[pos=0.25, above, sloped, font=\sffamily\tiny] {IMPROVE: targeted delta queries}
            ($(str.north)+(0.55,0.62)$) -- ($(str.north)+(0.55,0)$);

\node[wide, fill=blue!7, text width=154mm, align=center, anchor=south] (con) at (6.5,1.22)
  {\textbf{Conductor} (\texttt{main/SOUL.md}, Claude Code): fixed step order $\cdot$ goal and prior outputs passed to each
   subagent $\cdot$ file-persistence validation before every transition $\cdot$ checkpoint enforcement $\cdot$
   count re-derivation from decision logs};
\begin{scope}[on background layer]
  \node[band, fill=blue!5,   fit={(sco) (ga) ($(sco.north west)+(-0.1,0.42)$)}] (b1) {};
  \node[band, fill=green!6,  fit={(lim) (ana) ($(lim.north west)+(-0.1,0.42)$)}] (b2) {};
  \node[band, fill=violet!5, fit={(gb) (rev) ($(gb.north west)+(-0.35,0.42)$)}] (b3) {};
  \node[band, fill=black!4,  fit={(emb) (exp) (lg) ($(lg.north west)+(-0.25,0.42)$)}] (b4) {};
\end{scope}
\node[bandlab] at (b1.north west) {Layer 1 $\cdot$ Corpus construction};
\node[bandlab] at (b2.north west) {Layer 2 $\cdot$ Corpus analysis and quality};
\node[bandlab] at ($(b3.north west)+(1.75,0)$) {Layer 3 $\cdot$ Synthesis and review};
\node[bandlab] at (b4.north west) {Layer 4 $\cdot$ Export and maintenance};

\node[wide, fill=yellow!8, text width=154mm, anchor=north] (run) at (6.5,-6.55)
  {\textbf{Run folder: persisted shared state (one file per artifact)} --- \texttt{RUN.md} (tracker and conductor log),
   \texttt{scoper\_output}, \texttt{raw\_extraction\_list}, \texttt{borderline\_review}, \texttt{curation\_report},
   \texttt{master\_research\_data.csv}, \texttt{prisma\_analysis\_report.ipynb}, \texttt{quality\_report},
   \texttt{gap\_analysis}, \texttt{limits\_report}, figures and manifest, \texttt{references.bib},
   \texttt{synthesis\_outline}, \texttt{paper\_draft}, \texttt{review\_report}};
\coordinate (rwx) at (4.0,0);
\draw[{Stealth[length=1.8mm]}-{Stealth[length=1.8mm]}, semithick, black!60] (b4.south -| rwx) -- (run.north -| rwx)
  node[midway, right, font=\sffamily\tiny] {every agent reads and writes files; the Conductor validates them};

\node[wide, fill=black!3, text width=75mm, anchor=north west] (utl) at ($(run.south west)+(0,-0.15)$)
  {\textbf{Deterministic Python utilities} (\texttt{utils/}): \texttt{openalex\_utils}, \texttt{quality\_utils},
   \texttt{figures\_utils}, \texttt{api\_clients}, \texttt{pdf\_utils}, \texttt{csv\_utils}};
\node[wide, fill=black!3, text width=75mm, anchor=north east] (api) at ($(run.south east)+(0,-0.15)$)
  {\textbf{External services}: OpenAlex, Semantic Scholar, arXiv, Crossref, IEEE Xplore,
   ScienceDirect, Unpaywall, pandoc, pptxgenjs};
\end{tikzpicture}